\documentclass[twoside,11pt]{article}
\usepackage{jair, rawfonts}

\usepackage[utf8]{inputenc}
\usepackage{cite}
\usepackage[colorlinks]{hyperref}
\usepackage{url}
\usepackage{epsfig}
\usepackage{epstopdf}
\usepackage{graphicx}
\usepackage{amsmath}
\DeclareMathOperator*{\argmax}{arg\,max}
\usepackage{amssymb}
\usepackage{float}
\usepackage{breakcites}
\usepackage{booktabs}
\usepackage{forest}
\usepackage{rotating}
\usepackage{tcolorbox}
\usepackage{footnote}
\usepackage[T1]{fontenc}

\usepackage[fixed]{fontawesome5}

\usepackage{color, colortbl}
\definecolor{lightgray}{rgb}{0.90, 0.90, 0.90}
\usepackage{array}
\usepackage{multirow}

\newcommand\Tstrut{\rule{0pt}{2.6ex}}         
\newcommand\Bstrut{\rule[-0.9ex]{0pt}{0pt}}   

\newcolumntype{P}[1]{>{\centering\arraybackslash}p{#1}}

\newcommand{\etal}{\textit{et al.}}

\ShortHeadings{Explainable Deep Learning: \\
A Field Guide for the Uninitiated}{Ras, Xie, van Gerven, \& Doran}
\firstpageno{1}

\begin{document}

\title{Explainable Deep Learning: \\
A Field Guide for the Uninitiated}

\author{\name Gabri{\"e}lle Ras\footnotemark[1] \email g.ras@donders.ru.nl \\
      \addr Donders Institute for Brain, Cognition and Behaviour\\
      Radboud University Nijmegen\\
      6525 HR Nijmegen, the Netherlands 
      \AND
        \name Ning Xie\thanks{Gabri{\"e}lle Ras and Ning Xie are co-first authors on this work.}~\footnotemark[2] \email xining@amazon.com \\
       \addr 
       Amazon\\
       Seattle, WA 98109, USA
      \AND
       \name Marcel van Gerven \email m.vangerven@donders.ru.nl  \\
       \addr Donders Institute for Brain, Cognition and Behaviour\\
       Radboud University Nijmegen\\
       6525 HR Nijmegen, the Netherlands 
       \AND
       \name Derek Doran\thanks{Ning Xie and Derek Doran completed some of this work while at the Dept. of Computer Science and Engineering, Wright State University, Dayton, OH.}  \email derek.doran@tenet3.com \\
       \addr Tenet3, LLC \\
       Dayton, OH, USA
       }


\maketitle

\begin{abstract}
Deep neural networks (DNNs) have become a proven and indispensable machine learning tool. As a black-box model, it remains difficult to diagnose what aspects of the model's input drive the decisions of a DNN. In countless real-world domains, from legislation and law enforcement to healthcare, such diagnosis is essential to ensure that DNN decisions are driven by aspects appropriate in the context of its use. The development of methods and studies enabling  the explanation of a DNN's decisions has thus blossomed into an active, broad area of research. A practitioner wanting to study explainable deep learning may be intimidated by the plethora of orthogonal directions the field has taken. This complexity is further exacerbated by competing definitions of what it means ``to explain'' the actions of a DNN and to evaluate an approach's ``ability to explain''. This article offers a field guide to explore the space of explainable deep learning aimed at those uninitiated in the field. 
The field guide: 
i) Introduces three simple dimensions defining the space of foundational methods that contribute to explainable deep learning, 
ii) discusses the evaluations for model explanations,
iii) places explainability in the context of other related deep learning research areas,
and iv) finally elaborates on user-oriented explanation designing and potential future directions on explainable deep learning. 
We hope the guide is used as an easy-to-digest starting point for those just embarking on research in this field.
\end{abstract}

\section{Introduction}
\label{section:introduction}

Artificial intelligence (AI) systems powered by deep neural networks (DNNs) are pervasive across society: they run in our pockets on our cell phones~\cite{georgiev2017low}, in cars to help avoid car accidents~\cite{jain2015car}, in banks to manage our investments~\cite{chong2017deep} and evaluate loans~\cite{pham2017deep}, in hospitals to help doctors diagnose disease symptoms~\cite{nie2015disease}, at law enforcement agencies to help recover evidence from videos and images to help law enforcement~\cite{goswami2014mdlface}, in the military of many countries~\cite{lunden2016deep}, and at insurance agencies to evaluate coverage suitability and costs for clients~\cite{dong2016characterizing,sirignano2016deep}. However, when medical treatment is to be assigned or when a significant financial decision must be made, an AI that {\em suggests} a course of action with reasonable evidence, rather than to merely {\em prescribe} one, is desired. For the human ultimately responsible for the action taken, the use of DNNs leaves an important question unanswered: {\em how can a person that is held accountable for a decision trust a DNN's recommendation and justify its use?} Achieving trust and finding justification can hardly be achieved if the user does not have access to a satisfactory explanation for the process that led to the recommendation. Consider, for example, a hypothetical scenario in which a medical system runs a DNN in the backend. Assume that the system makes life-altering predictions about whether or not a patient has a terminal illness. It is desirable if this system could also provide a rationale behind its predictions. Equally important is for the system to give a rationale that both physicians and patients can understand and trust. Trust in a decision is built upon a rationale that is: (i) easily interpretable; (ii) relatable to the user; (iii) connects the decision with contextual information about the choice or to the user's prior experiences; and (iv) reflects the intermediate thinking of the user in reaching a decision. Given the qualitative nature of these characteristics, it may come as no surprise that there is great diversity in the definitions, approaches, and techniques used by researchers to provide a rationale for the decisions of a DNN. This diversity is further compounded by the fact that
the form of a rationale often conforms to a researcher's personal notion of what constitutes an ``explanation''. For a newcomer to the field, whether a seasoned researcher or students in a discipline that DNN's are impacting, jumping into the field is a daunting task. 

\begin{figure}
    \centering
    \includegraphics[width=0.9\linewidth]{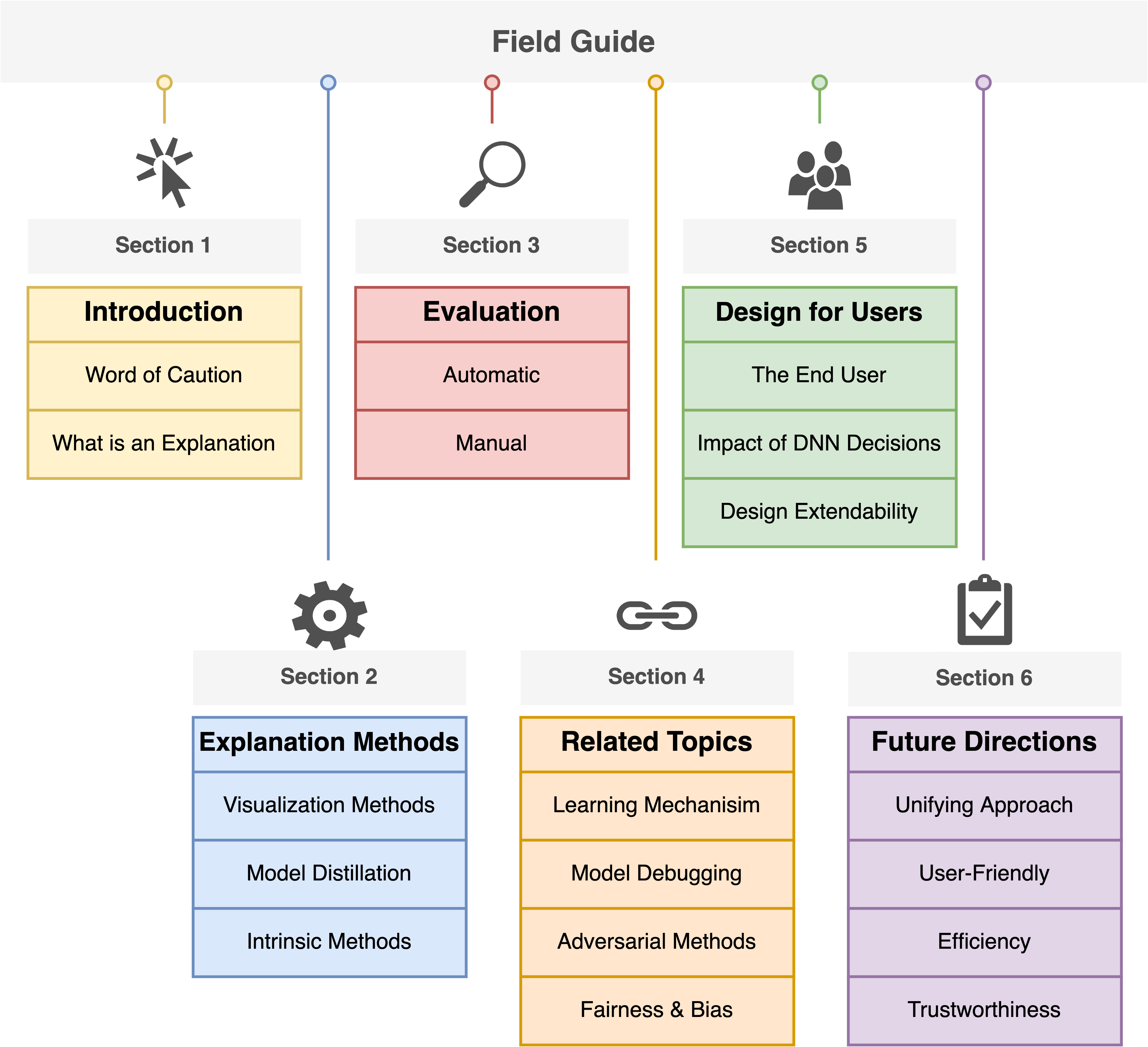}
    \caption{Outline of the field guide.}
    \label{fig:outline}
\end{figure}

This article offers a starting point for researchers and practitioners who are
embarking into the field of explainable deep learning. This field guide is designed to help an uninitiated researcher understand:

\begin{itemize}
    \item A set of {\bf dimensions} characterizing the space of foundational work in explainable deep learning, and a description of such methods. 
    This space summarizes the core aspects of explainable DNN techniques that a majority of present work is inspired by or built from (\textbf{Section~\ref{section:methods}}).
    \item Methods for \textbf{evaluating} explanation methods (\textbf{Section~\ref{section:evaluation}}).
    \item Complementary research {\bf topics} that are aligned with explainability. 
    Topics that are complementary to explainability may involve the development of mechanisms that mathematically explain how DNNs learn to generalize, or approaches to reduce a DNN's sensitivity to particular input features. 
    Such topics are indirectly associated with explainability in the sense that they investigate {\em how} a DNN learns or performs inference, even though the intention of the work is not directly to investigate explanations (\textbf{Section~\ref{section:topics_associated}}).
    \item The {\bf considerations} of a designer developing an explainable DNN system (\textbf{Section~\ref{section:designing}}).
    \item {\bf Future directions} in explainability research (\textbf{Section~\ref{section:future}}).  
\end{itemize}

Our taxonomy of explainable DNN techniques clarifies the technical ideas underpinning most modern explainable deep learning techniques. The discussion of fundamental explainable deep learning methods, emblematic of each framework dimension, provides further context for the modern work that builds on or takes inspiration from them. Following the taxonomy is a brief discussion on the evaluations of model explanations. Complementary DNN topics are reviewed, and the relationships between explainable DNNs and other related research areas are developed. The field guide then turns to essential considerations that need to be made when building an explainable DNN system in practice, considering the end-user. Finally, the overview of our current limitations and seldom-looked at aspects of explainable deep learning suggests new research directions. This information captures what a newcomer needs to know to successfully navigate the current research literature on explainable deep learning and identify new research problems.

There are many reviews on the topic of model explainability. Most of them focus on explanations of general artificial intelligence methods~\cite{arrieta2019explainable,carvalho2019machine,mueller2019explanation,tjoa2019survey,gilpin2018explaining,adadi2018peeking,miller2018explanation,guidotti2018survey,lipton2018mythos,liu2017towards,dovsilovic2018explainable,doshi2017towards,molnar2020interpretable,carvalho2019machine}, and some on deep learning~\cite{ras2018explanation,montavon2018methods,zhang2018visual,samek2017explainable,erhan2010understanding}. 

The unique contributions of this field guide are as follows. First, it specifically targets explanations for deep learning systems, while existing reviews focus on explanations of general artificial intelligence methods or have a narrower focus on particular types of deep learning architectures. Second, it specifically targets researchers uninitiated in explainable deep learning and aims to lower the bar to enter this field. Third, it introduces a novel categorization scheme to systematically organize numerous explanation methods, with an eye towards simplicity and focus on the field's foundations. Finally, it elaborates on other topics closely related to explainable deep learning. The review connects related fields to explainable deep learning to better understand how they contribute to existing work to improve DNN transparency, robustness, and reliability. 

\subsection{A Word of Caution}
Although this paper focuses on explaining DNNs, it does not mean that DNNs are the only problem-solving tools in a machine learning toolbox. DNNs have significant potential for misuse when applied prematurely or incorrectly. Extreme caution must be taken, by all parties involved, to ensure that the DNN technology and derivatives are properly tested before production and commercial use. It is also recommended to review the perceived need for DNNs and consider if other algorithms can serve the same purpose~\cite{rudin2019stop}. One recent example of this is the wrongful use of facial recognition technology by federal law enforcement agencies. Studies have shown that facial recognition technology is significantly less accurate on non-white, non-male persons~\cite{buolamwini2018gender,garvie2016perpetual}, yet the immature technology was used in the real world with negative outcomes. 

On the other hand, DNN applications in domains such as medicine~\cite{rajpurkar2018deep,esteva2017dermatologist} or applications that can benefit climate change~\cite{rolnick2019tackling} have been meaningful and beneficial to society as a whole. This paper looks at explainability through the lens of the DNNs. However, that does not mean that explanations are only required for DNNs. The need for explanations extend to the whole of scientific reasoning.

\subsection{But What is an Explanation?}

\begin{figure}
    \centering
    \includegraphics[width=0.9\linewidth]{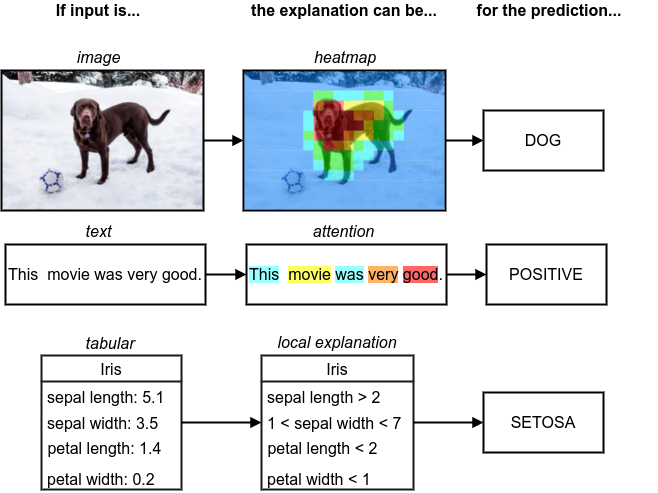}
    \caption{Examples of what explanations can look like in practice. The explanation depends on the type of data used and the method used to create the explanation. In general an explanation is any information that aids the user in understanding the model rationale behind the model prediction.}
    \label{fig:what_is_explanation}
\end{figure}

In general, defining \textit{explanation} is a philosophical activity that, while worth undertaking, does not align with the goal of this paper. Instead, we will give examples of what explanations can look like in the context of deep learning.  
In its most general form, an explanation is any information that can help the user understand and communicate to others why the model exhibits a particular pattern of decision-making and how individual decisions come about. 

The goal of any explanation can roughly fall into one of the following two categories: (i) {\em explanations that give insight into model training and generalization.} These explanations give a practitioner additional information that can be used to make decisions about the components in the model training and validation process, e.g., the number of labeled data, value of the hyperparameters, model choice. The other category is (ii) {\em explanations that give insight into model predictions.} Most explanations fall into this category and help practitioners explain why the model made a particular prediction, usually in terms of the model input. These explanations can be used to communicate to others (potentially non-experts) about model predictions. Many individual predictions can be analyzed to reveal patterns in overall model prediction behavior. This category of explanations can further be broken down in more specific categories such as \textit{counterfactual explanations} \cite{verma2020counterfactual} and \textit{contrastive explanations} \cite{miller2018contrastive}. 

Most explanations bear a strong resemblance to the data type that was used to train the DNN. If the datatype is an image, the explanation can be a saliency or heatmap. A saliency map depicts regions in the image that the explanation method determined was important for the network's prediction. If the datatype is text-based, the explanation can look like highlighted words in the text. The explanation method (for instance, attention visualization) determines which words are highlighted. If the data is composed of attributes, i.e., data that can be represented as a table, the explanation can be a set of rules that describe which combinations of different attribute values lead to which predictions. The illustrations in Figure~\ref{fig:what_is_explanation} reflect different explanation representations based on data type. It is useful to keep in mind that while heatmaps are one of the most common and natural ways to present a model explanation, it is also subject to interpretation by the practitioner. Practitioners can interpret the explanation differently, especially when the explanation method is unknown.

\section{Methods for Explaining DNNs}
\label{section:methods}

\begin{figure}
    \begin{tabular}{c}
    \resizebox{0.80\columnwidth}{!}{\begin{forest} 
    for tree={
    edge path={
          \noexpand\path[\forestoption{edge}](!u.parent anchor) -- +(20pt,0) |- (.child anchor)\forestoption{edge label};},
    l sep=5em, s sep=1em,
    child anchor=west,
    parent anchor=east,
    grow'=0,
    line width=0.75mm,
    anchor=west,
    draw,
    align=left,
    }
    [{{\Huge Explaining}}\\{{\Huge DNNs}}\\{{\Huge Methods}}
    [{{\Huge Visualization}}
    [{{\huge Backpropagation}}
    [{{\huge \textit{Activation Maximization}}} \\ 
    {{\huge \cite{erhan2009visualizing}}}
    ]
    [{{\huge \textit{Deconvolution}}} \\ 
    {{\huge \cite{zeiler2011adaptive};\cite{zeiler14}}}
    ]
    [{{\huge \textit{CAM and Grad-CAM}}} \\
    {{\huge \cite{zhou2016learning};}}
    {{\huge \cite{selvaraju2017grad}}}
    ]
    [{{\huge \textit{LRP}}} \\                         
    {{\huge \cite{bach15};}}
    {{\huge \cite{lapuschkin2016analyzing}}}\\
    {{\huge \cite{arras2016explaining};}}
    {{\huge \cite{arras2017relevant}}}\\
    {{\huge \cite{ding2017visualizing};}}
    {{\huge \cite{montavon2017explaining}}}
    ]
    [{{\huge \textit{DeepLIFT}}} \\                         
    {{\huge \cite{shrikumar2017learning}}}
    ]
    [{{\huge \textit{Integrated Gradients}}} \\      
    {{\huge \cite{sundararajan2016gradients,sundararajan2017axiomatic}}}
    ]
    ]
    [{{\huge Perturbation}}
    [{{\huge \textit{Occlusion Sensitivity}}} \\ 
    {{\huge \cite{zeiler14}; \cite{zhou2014object}}}\\
    ]
    [{{\huge \textit{Representation Erasure}}} \\                         
    {{\huge \cite{li2016understanding}}}
    ]
    [{{\huge \textit{Meaningful Perturbation}}} \\                         
    {{\huge \cite{fong2017interpretable}}}\\
    ]
    [{{\huge \textit{Prediction Difference Analysis}}} \\  
    {{\huge \cite{zintgraf2017visualizing}}}\\
    {{\huge \cite{robnik2008explaining}}}\\
    ]
    ]
    ]  
    [
    {{\Huge Distillation}} 
    [{{\huge Local Approximation}}
    [{{\huge \textit{LIME}}} \\                         
    {{\huge \cite{ribeiro2016should,ribeiro2016model}}}
    ]
    [{{\huge \textit{Anchor-LIME}}} \\ 
    {{\huge \cite{ribeiro2016nothing}}}
    ]
    [{{\huge \textit{Anchors}}} \\ 
    {{\huge \cite{ribeiro2018anchors}}}
    ]
    [{{\huge \textit{STREAK}}} \\ 
    {{\huge \cite{elenberg2017streaming}}}
    ]
    [{{\huge \textit{SHAP}}} \\ 
    {{\huge \cite{lundberg2017unified}}}
    ]
    [{{\huge \textit{Causal SHAP}}} \\ 
    {{\huge \cite{heskes2020causal}}}
    ]
    ]
    [{{\huge Model  Translation}} 
    [{{\huge \textit{Tree Based}}} \\   
    {{\huge \cite{frosst2017distilling}}}\\
    {{\huge \cite{tan2018learning};}}
    {{\huge \cite{zhang2019interpreting}}}\\
    ]
    [{{\huge \textit{FSA Based}}} \\            
    {{\huge \cite{hou2020learning}}}
    ]
    [{{\huge \textit{Graph Based}}} \\            
    {{\huge \cite{zhang2017growing,zhang2017interpreting}}}
    ]
    [{{\huge \textit{Rule Based}}} \\ 
    {{\huge \cite{murdoch2017automatic};}}
    {{\huge \cite{harradon2018causal}}}
    ]
    ]
    ]
    [
    {{\Huge Intrinsic}}
    [{{\huge  Attention Mechanisms }} \\
    [{{\huge \textit{Single-Modal Weighting}}} \\ 
    {{\huge \cite{bahdanau2014neural};}}
    {{\huge \cite{luong2015effective}}}\\
    {{\huge \cite{wang2016attention};}}
    {{\huge \cite{vaswani2017attention}}}\\
    {{\huge \cite{letarte2018importance};}}
    {{\huge \cite{he2018effective}}}\\
    {{\huge \cite{devlin2019bert}}}
    ]
    [{{\huge \textit{Multi-Modal Interaction}}} \\ 
    {{\huge \cite{vinyals2015show};}}
    {{\huge \cite{xu2015show}}} \\
    {{\huge \cite{antol2015vqa};}}
    {{\huge \cite{park2016attentive}}}\\
    {{\huge \cite{goyal2017making};}}
    {{\huge \cite{teney2018tips}}}\\
    {{\huge \cite{mascharka2018transparency};}}
    {{\huge \cite{anderson2018bottom}}}\\
    {{\huge \cite{xie2019visual}}}\\
    ]
    ]
    [{{\huge  Joint Training}} \\        
    [{{\huge \textit{Text Explanation}}} \\ 
    {{\huge \cite{hendricks2016generating};}}
    {{\huge \cite{camburu2018snli}}}\\
    {{\huge \cite{park2018multimodal};}}
    {{\huge \cite{kim2018textual}}}\\
    {{\huge \cite{zellers2019recognition};}}
    {{\huge \cite{liu2019towards};}}
    {{\huge \cite{hind2019ted}}}
    ]
    [{{\huge \textit{Explanation Association}}} \\ 
    {{\huge \cite{lei2016rationalizing};}} 
    {{\huge \cite{dong2017improving}}}\\ 
    {{\huge \cite{melis2018towards};}} 
    {{\huge \cite{iyer2018transparency}}}\\ 
    ]
    [{{\huge \textit{Model Prototype}}} \\ 
    {{\huge \cite{li2018deep}; \cite{chen2019looks}}}
    ]
    ]
    ]
    ]
    \end{forest}} 
    \end{tabular}
    \caption{Methods for explaining DNNs.}
    \label{figure:methods}
\end{figure}
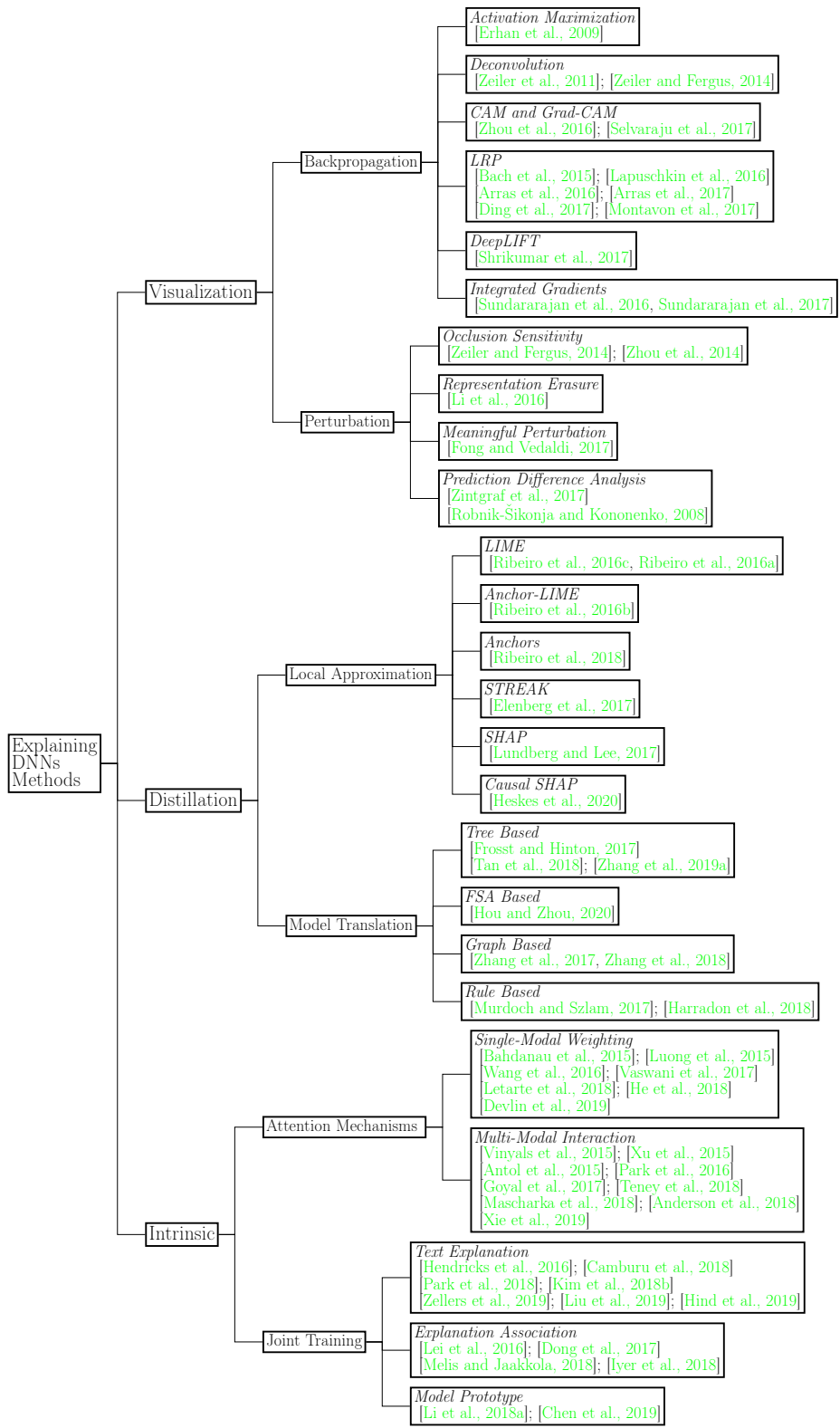

There are countless surveys on explainable AI~\cite{arrieta2019explainable,carvalho2019machine,mueller2019explanation,tjoa2019survey,gilpin2018explaining,adadi2018peeking,miller2018explanation,guidotti2018survey,lipton2018mythos,liu2017towards,dovsilovic2018explainable,doshi2017towards} and explainable deep learning~\cite{ras2018explanation,montavon2018methods,zhang2018visual,samek2017explainable,erhan2010understanding}. The many surveys cover a large body of work that may prove hard to navigate and synthesize into a broad view of the field. Instead, this article investigates a simple space of foundational explainable DNN methods. We say a method is foundational if it is often used in practice or if it introduces a concept that modern work builds upon. Understanding this smaller space of foundational methods will support a reader as they study modern approaches. 

Since different users at different stages of the software pipeline have different requirements, it is only possible to represent relative advantages given the explainability goal that needs to be achieved. The users that we will take into account for this discussion are the \textit{expert users} as described in \cite{ras2018explanation}. 

We present a simple three-dimensional space encompassing: 

\begin{itemize}
    \item \textbf{Visualization methods}: Visualization methods express an explanation by highlighting, through a scientific visualization, characteristics of an input that strongly influence the output of a DNN. 
    \item \textbf{Model distillation}: Model distillation develops a separate, ``white-box'' machine learning model that is trained to mimic the input-output behavior of the DNN. The white-box model, which is inherently explainable, is meant to identify the decision rules or input features influencing DNN outputs. 
    \item \textbf{Intrinsic methods}: Intrinsic methods are DNNs that have been specifically created to render an explanation along with its output. 
    As a consequence of its design, intrinsically explainable deep networks can jointly optimize  model performance and a quality of the explanations produced. 
\end{itemize}

\subsection{Visualization Methods}
\label{subsection:attribution_methods}

\begin{table}[ht]
    \centering
    \resizebox{\textwidth}{!}{
    \begin{tabular}{p{5cm}p{6cm}p{8cm}}
    \specialrule{.2em}{.1em}{.1em}
    \textbf{Visualization Methods} & \textbf{Summary} & \textbf{References} \\ 
    \specialrule{.1em}{.1em}{.1em}
    Backpropagation-based &  Visualize feature relevance from volume of gradient 
    passed through layers during network training. & \cite{erhan2009visualizing,zeiler2011adaptive,zeiler14,zhou2016learning,selvaraju2017grad,bach15,lapuschkin2016analyzing,arras2016explaining,arras2017relevant,ding2017visualizing,montavon2017explaining,shrikumar2017learning,sundararajan2017axiomatic,sundararajan2016gradients} \\ \hline
    
    Perturbation-based & Visualize feature relevance by comparing network output of an input and a
    modified copy of the input.  & \cite{zeiler14,zhou2014object,li2016understanding,fong2017interpretable,robnik2008explaining,zintgraf2017visualizing} \\ 
    
    \specialrule{.2em}{.1em}{.1em}
    \end{tabular}
    }
    \caption{Visualization methods.}
    \label{table:visualization_methods}
\end{table}

Visualization methods associate the degree to which a DNN considers input features to a decision. This association is often referred to as {\em attribution}. A common explanatory form of visualization methods is {\em saliency maps} or {\em heatmaps}, where oftentimes a transparent colored heatmap is overlaid on the original input image. These maps identify input features that are most salient, in the sense that they cause a maximum response or stimulation influencing the model's output~\cite{yosinski2015understanding,uozbulak2019pytorch,olah2017feature,olah2018the,carter2019activation}. 
We break down visualization methods into two types, namely {\em backpropagation} and {\em perturbation}-based visualization.
The types are summarized in Table~\ref{table:visualization_methods} and will be discussed further below. 

\begin{figure}[ht]
    \centering
    \includegraphics[width=0.95\linewidth]{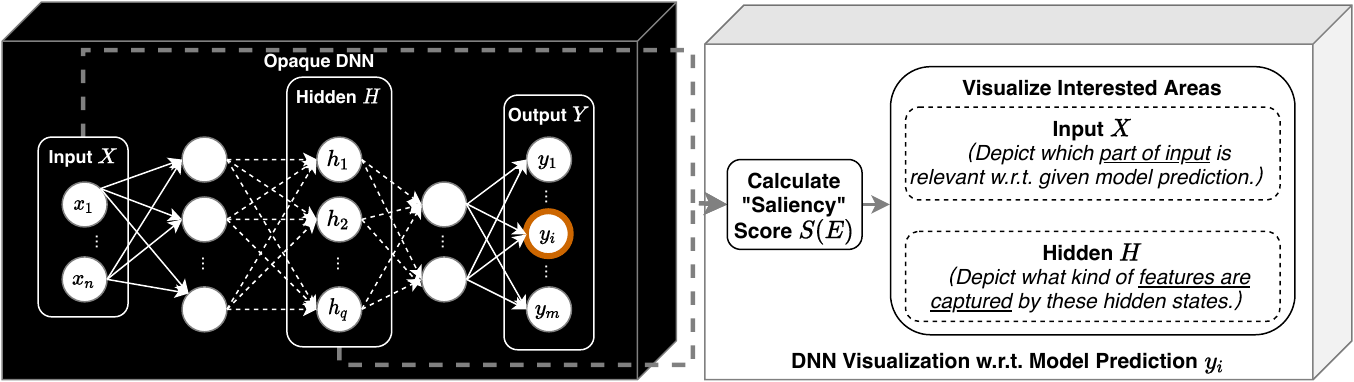}
    \caption{Visualization Methods. The to-be-visualized element $E$ can be from either the model input $X$ or hidden states $H$. Visualization is based on the calculated saliency score $S(E)$, which varies along with different visualization methods.}
    \label{fig:visualization}
\end{figure}

\subsubsection{Backpropagation-based methods}
Backpropagation-based methods identify the saliency of input features based on some evaluation of gradient signals passed from output to input during network training. A baseline gradient-based approach visualizes the partial derivative of the network output with respect to each input feature scaled by its value~\cite{simonyan2013deep,springenberg2014striving}, thus quantifying the sensitivity of the network's output with respect to input features. In a scene recognition task, for example, a high relevance score for pixels representing a bed in a CNN that decides the image is of class ``bedroom'' may suggest that the decision made by the CNN is highly sensitive to the presence of the bed in the image. Other gradient-based methods may evaluate this sensitivity with respect to the output, but from different collections of feature maps at intermediate CNN network layers~\cite{zeiler14,bach15,montavon2017explaining,shrikumar2017learning}.
\newline

\noindent \textbf{Activation maximization.} One of the earliest works on visualization in deep architectures is proposed by~\cite{erhan2009visualizing}. This seminal study introduces the activation maximization method to visualize important features in any layer of a deep architecture by optimizing the input $\boldsymbol{X}$ such that the activation $a$ of the chosen unit $i$ in a layer $j$ is maximized:

\begin{equation}
    \argmax_{\boldsymbol{X}} a_{i, j}(\boldsymbol{X}, \boldsymbol{\theta})
\end{equation}

Parameters $\boldsymbol{\theta}$ of a trained network are kept fixed during activation maximization. The optimal $\boldsymbol{X}$ is found by computing the gradient of $a_{i,j}(\boldsymbol{X}, \boldsymbol{\theta})$ and updating $\boldsymbol{X}$ in the direction of the gradient. The practitioner decides the values of the hyperparameters for this procedure, i.e., the learning rate and how many iterations to run. The optimized $\boldsymbol{X}$ will be a representation, in the input space, of the features that maximize the activation of a specific unit, or if the practitioner chooses so, multiple units in a specific network layer. 

By visualizing the internal representations, the practitioner can check if concepts learned by the model are human interpretable. The quality of the concepts can be used as an indication of model generalizability and determine if, for instance, additional labeled data is needed to train the model. Activation maximization can give insight into the model training and generalization process but does not lend itself to explaining individual model predictions.
\newline

\noindent \textbf{Deconvolution.} 
Deconvolution was originally introduced as an algorithm to learn image features in an unsupervised manner~\cite{zeiler2011adaptive}. However, the method gained popularity because of its applications in visualizing higher layer features in the input space~\cite{zeiler14}, i.e., visualizing higher layer features in terms of the input. Deconvolution assumes that the model being explained is a neural network consisting of multiple convolutional layers. We will refer to this model as $\texttt{CNN}$. The consecutive layers of this network consist of a convolution of the previous layer's output (or the input image in the case of the first convolutional layer) with a set of learned convolutional filters, followed by the application of the rectified linear function (ReLU) $\texttt{ReLU}(\boldsymbol{A}) = \max(\boldsymbol{A}, 0)$ on the output of the aforementioned convolution

\begin{equation}
    \label{eq:conv_layer}
    \boldsymbol{A}^{\ell}, s^\ell = \texttt{maxpool}\left(\texttt{ReLU}\left(\boldsymbol{A}^{\ell-1} * \boldsymbol{K}^{\ell} + \boldsymbol{b}^{\ell}\right)\right)
\end{equation}

where $\ell$ indicates the respective layer, $\boldsymbol{A}^\ell$ is the output of the previous layer, $\boldsymbol{K}$ is the learned filter, and $\boldsymbol{b}$ is the bias. If the outputs from the ReLU are passed through a local max-pooling function, it additionally stores the output $s^\ell$ containing the indices of the maximum values for a later unpooling operation. In the original paper, the set of $s^\ell$'s are referred to as \textit{switches}. A deconvolutional neural network, referred to as $\texttt{DeCNN}$, consists of the inverse operations of the original convolutional network $\texttt{CNN}$. $\texttt{DeCNN}$ takes the output of $\texttt{CNN}$ as its input. In other words, $\texttt{DeCNN}$ runs the $\texttt{CNN}$ in reverse, from top-down. This is why the deconvolution method is classified as a backpropagation method. The convolutional layers in $\texttt{CNN}$ are replaced with \textit{deconvolutions} and the max-pooling layers are replaced with \textit{unpooling} layers. A deconvolution is also called a \textit{transposed convolution}, meaning that the values of $\boldsymbol{K}^\ell$ are transposed and then copied to the deconvolution filters ${\boldsymbol{K}^\ell}^T$. If the $\texttt{CNN}$ included max-pooling layers, they are replaced with unpooling layers which approximately upscales the feature map, retaining only the maximum values. This is done by retrieving the indices stored in $s^\ell$ at which the maximum values were located when the max-pooling was originally applied in $\texttt{CNN}$. 

As an example let us see the calculations involved in deconvolving Equation \ref{eq:conv_layer}:

\begin{equation}
    \boldsymbol{A}^{\ell-1} = \texttt{unpool}\left(\texttt{ReLU}\left(\left(\boldsymbol{A}^{\ell} - \boldsymbol{b}^{\ell} \right) * {\boldsymbol{K}^{\ell}}^T \right), s^{\ell}\right)
    \label{eq:deconv}
\end{equation}

Using Equation \ref{eq:deconv} one or multiple learned filters $\boldsymbol{K}$ in any layer of the network can be visualized by reverse propagating the values of $\boldsymbol{K}$ all the way back to the input space. Finally, this study also describes how the visualizations can be used for architecture selection. 

Practitioners can use this method to visualize how much information from the original input the extracted features retain, and gain insight on how information is extracted from data at different network layers. From this insight actions can be taken to improve the model training process. Deconvolution does not lend itself to explaining single model predictions. 
\newline

\noindent \textbf{CAM and Grad-CAM.} \cite{zhou2016learning} describes a visualization method for creating class activation maps (CAM) using global average pooling (GAP) in CNNs. \cite{lin2013network} proposes the idea to apply a global average pooling on the activation maps of the last convolutional layer, right before the fully connected (FC) output layer. This results in the following configuration at the end of the CNN: $\texttt{GAP(Conv)} \rightarrow \texttt{FC} \rightarrow \texttt{softmax}$. The FC layer has $C$ nodes, one for each class. The CAM method combines the activations $\boldsymbol{A}$ from \texttt{Conv}, containing $K$ convolutional filters, and weights $w_{k,c}$ from \texttt{FC}, where the $(k, c)$ pair indicates the specific weighted connection from \texttt{Conv} to \texttt{FC}, to create relevance score map:

\begin{equation}
    map_c = \sum_k^K w_{k,c} \boldsymbol{A}_k
\end{equation}

The map is then upsampled to the size of the input image and overlaid on the input image, very similar to a heat map, resulting in the class activation map. Each class has a unique CAM, indicating the image regions  important to network prediction for that class. CAM can only be applied on CNNs that employ the $\texttt{GAP(Conv)} \rightarrow \texttt{FC} \rightarrow \texttt{softmax}$ configuration.

Gradient-weighted Class Activation Map (\textbf{Grad-CAM})~\cite{selvaraju2017grad} is a generalization CAM using the gradients of the network output with respect to the last convolutional layer to achieve the class activation map. 
This allows Grad-CAM to be applicable to a broader range of CNNs compared to CAM, only requiring that the final activation function used for network prediction to be a differentiable function, e.g., softmax. For each feature map $\boldsymbol{A}_k$ in the final convolutional layer of the network, a gradient of the score $y_c$ (the value before softmax, also known as \textit{logit}) of class $c$ with respect to every node in $\boldsymbol{A}_k$ is computed and averaged to get an importance score $\alpha_{k,c}$ for feature map $\boldsymbol{A}_k$:

\begin{equation}
    \alpha_{k,c} = \frac{1}{m\cdot n}\sum_{i=1}^m\sum_{j=1}^n \frac{\partial y_c}{\partial \boldsymbol{A}_{k,i,j}}
    \label{eq:gradcam}
\end{equation}

where $\boldsymbol{A}_{k,i,j}$ is a neuron positioned at $(i, j)$ in the $m \times n$ feature map $\boldsymbol{A}_k$. Grad-CAM linearly combines the importance scores of each feature map and passes them through a ReLU to obtain an $m \times n$-dimensional relevance score map

\begin{equation}
    \label{eq:relu}
    map_c = \texttt{ReLU}\left(\sum_k^K \alpha_{k,c} \boldsymbol{A}_k \right)
\end{equation}

\begin{figure}[ht]
    \centering
    \includegraphics[width=0.90\linewidth]{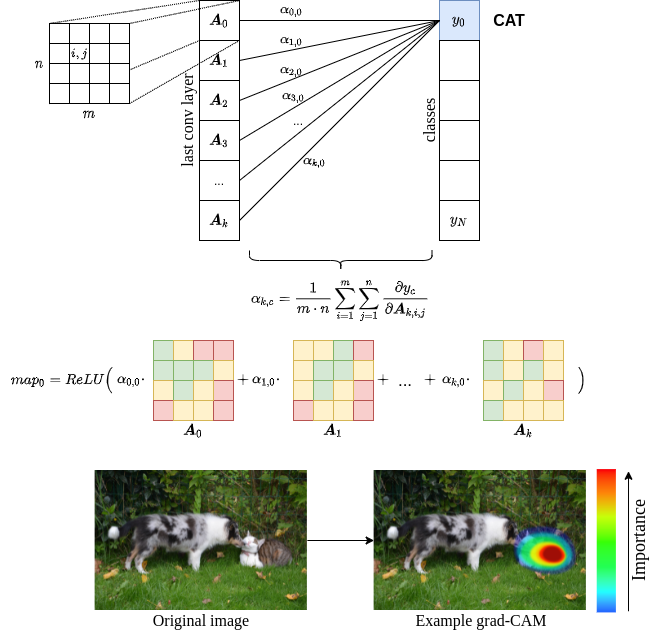}
    \caption{Visual explanation of how grad-CAM works. \textbf{Top}: Visualization of Equation \ref{eq:gradcam} for calculating the importance scores $\alpha_{i, j}$ for each feature map $\boldsymbol{A}_k$. \textbf{Middle}: The heatmap for a specific class is computed by multiplying the importance score with each feature map and taking the sum. Afterwards the heatmap is upsampled and overlaid on the original image. \textbf{Bottom}: Heatmap for the prediction ``cat''.}
    \label{fig:grad-CAm}
\end{figure}

The relevance score map is then upsampled via bilinear interpolation to be of the same dimension as the input image to produce the class activation map. Figure \ref{fig:grad-CAm} shows a visual representation of the grad-CAM method and an example of what a grad-CAM heatmap looks like for the prediction ``cat''. 

Practitioners can use the CAM family of methods to determine, given an input and a class, what is the information in the input that gives evidence for that class. Based on this information the practitioner can determine to what extent model predictions can be interpreted and assess for which classes consistent model predictions can expected. For example, if we have two models where both models have the same accuracy score, a model that produces heatmaps consistent with human experience is often considered more trustworthy compared to one where the heatmaps correspond poorly to human experience. Practitioners can also use the CAM family of methods to determine if there is an unfavorable class bias that the model is picking up on, e.g., skin color. 
The reader should note that only positive attribution can be computed with this method due to the ReLU function in Equation \ref{eq:relu}.
\newline

\noindent \textbf{Layer-Wise Relevance Propagation.}
LRP methods create a saliency map that, rather than measuring sensitivity, 
represents the relevance of each input feature to the output of the network~\cite{bach15,lapuschkin2016analyzing,arras2016explaining,arras2017relevant,ding2017visualizing,montavon2017explaining}. While \textit{sensitivity} measures the change in response in the network's output as a result of changing attributes in the input~\cite{kindermans2019reliability}, \textit{relevance} measures the strength of the connection between the input features to   network output, without making any changes to the input or the components of the network. LRP methods decompose the output value $f(\boldsymbol{x})$ of a deep network $f$ across input features $\boldsymbol{x} = (x_1, x_2, \ldots, x_N)$, such that $f(\boldsymbol{x}) = \sum_i r_i$ where $r_i$ is the relevance score of feature $x_i$. 

Perhaps the most generic type of LRP is called Deep Taylor Decomposition~\cite{montavon2017explaining}. The method is based on the fact that $f$ is differentiable and hence can be approximated by a Taylor expansion of $f$ at some root $\hat{\boldsymbol{x}}$ for which $f(\hat{\boldsymbol{x}}) = 0$:

\begin{equation} \label{}
    \begin{split}
    f(\boldsymbol{x}) &= f(\hat{\boldsymbol{x}}) + \nabla_{\hat{\boldsymbol{x}}}f \cdot (\boldsymbol{x} - \hat{\boldsymbol{x}}) + \epsilon \\
    & =  \sum_i^N \frac{\partial f}{\partial x_i}(\hat{x_i}) \cdot (x_i - \hat{x_i}) + \epsilon        
    \end{split}
\end{equation}

where $\epsilon$ encapsulates all second order and higher terms
in the Taylor expansion. A good root point is one that is as minimally different from $\boldsymbol{x}$ and that causes the function $f(\boldsymbol{x})$ to output a different prediction. The relevance score for inputs can then be seen as the terms inside of the summation:

\begin{equation}
    r_i = \frac{\partial f}{\partial x_i}(\hat{x_i}) \cdot (x_i - \hat{x_i})  
\end{equation}

To extend this idea to a DNN, the deep Taylor decomposition algorithm considers a conservative decomposition of relevance scores across layers of the network, starting from the output, through each hidden layer, back to the input. Thus, the method requires that the relevance score of a node $i$ at layer $l$, denoted $r_i^l$ be decomposable into

\begin{equation}
    r_i^\ell = \sum_j^{M} r_{i,j}^\ell  
\end{equation}

where the summation is taken over all $M$ nodes in layer $\ell+1$ that node $i$ in layer $\ell$ connects or contributes to. This indicates that the relevance score of the later layers can be backpropagated to generate the relevance score of former layers. The relevance score with respect to the input space can thus be calculated by conducting this decomposition rule layer by layer. Further details can be found in the original paper~\cite{montavon2017explaining}. 

In practice, the results of applying LRP are similar to the results of the CAM family of methods: given an input and a prediction, both methods tell the practitioner the regions in the input that are most relevant for the prediction. LRP allows the heatmap to display negative attributions in addition to positive ones. Practitioners can use the information to assess and further investigate things like model bias, prediction consistency and model trust. However, with LRP the practitioner has to additionally supply the method with a reference input (root) $\hat{\boldsymbol{x}}$, which in some cases can either be unsolvable or expensive to compute. It is worth noting that the visualization results of LRP rely on the root choices: depending on the input space restrictions, a different root can be chosen, and for different roots chosen, the relevance propagation rule varies, which ultimately yields different appearances of the heatmap. Compared to CAM methods, the LRP heatmap could be of higher quality and more precise. This is because LRP assigns each individual pixel a relevance score, as opposed to CAM, which looks at activation maps in the final layer. The final CAM heatmap will be an upsampled image indicating an approximate relevant region.
\newline

\noindent \textbf{DeepLIFT.} 
Deep Learning Important FeaTures (DeepLIFT) is an important approach based on backpropagation  by~\cite{shrikumar2017learning}. It assigns relevance scores to input features based on the difference between an input $\boldsymbol{x}$ and a ``reference'' input $\boldsymbol{x}'$. The reference should be chosen according to the problem at hand and can be found by answering the question \textit{``What am I interested in measuring differences against?''}. In an example using MNIST the reference chosen is an input of all zeros as this is the background value in the images. Define $\Delta t = f(\boldsymbol{x}) - f(\boldsymbol{x}')$ as the difference-from-reference of an interested neuron output of the network between $\boldsymbol{x}$ and reference $\boldsymbol{x}'$, and $\Delta \boldsymbol{x} = \boldsymbol{x} - \boldsymbol{x}'$ as the difference between $\boldsymbol{x}$ and $\boldsymbol{x}'$. DeepLIFT assigns a relevance score $R_{\Delta x_i \Delta t} $ for input feature $x_i$: 

\begin{equation}
    \label{eq:delta_t}
    \Delta t = \sum_{i=1}^N R_{\Delta x_i \Delta t}
\end{equation}

where $N$ is the number of input neurons that are necessary to compute $t$.
In this formulation, $R_{\Delta x_i \Delta t}$ can be thought of as a weight denoting how much influence $\Delta x_i$ had on $\Delta t$. According to Equation \ref{eq:delta_t} the sum of the all weights is equal to the difference-from-reference output $\Delta t$. The relevance score can be calculated via the \textit{Linear} rule, \textit{Rescale} rule, or \textit{RevealCancel} rule, as elaborated in their study. A multiplier $m_{\Delta \boldsymbol{x} \Delta t}$ is defined as

\begin{equation}
    m_{\Delta \boldsymbol{x} \Delta t} = \frac{R_{\Delta \boldsymbol{x} \Delta t}}{\Delta \boldsymbol{x}}
\end{equation}

indicating the relevance of $\Delta \boldsymbol{x}$ with respect to $\Delta t$, averaged by $\Delta \boldsymbol{x}$. Given a hidden layer $\ell$ of nodes $\boldsymbol{a}^\ell = (a^{\ell}_1, a^{\ell}_2, \ldots a^{\ell}_K)$, whose upstream connections are the input nodes $\boldsymbol{x} = (x_1, x_2, \ldots x_N)$, and a downstream target node is $t$, the DeepLIFT paper proves the effectiveness of the ``chain rule'' as illustrated below:

\begin{equation}
    m_{\Delta x_i \Delta t} = \sum_{j=1}^K m_{\Delta x_i \Delta a^\ell_j} m_{\Delta a^\ell_j \Delta t}  
\end{equation}

This ``chain rule'' allows for layer-by-layer computation of the relevance scores of each hidden layer node via backpropagation. The DeepLIFT paper and appendix specify particular rules for computing $m_{\Delta x_i \Delta a^\ell_j}$ based on the architecture of the hidden layer $\boldsymbol{a}^\ell$.

DeepLIFT resembles LRP because the practitioner has to choose a reference point and because the algorithm assigns pixel-wise relevance scores. The heatmap results of DeepLIFT explain the difference in prediction between the reference image and the original prediction. Heatmap interpretation depends on which reference image is chosen. By being creative with the choice of the reference image, the practitioner can use DeepLIFT to probe the network in different ways. Heatmaps obtained with DeepLIFT can also display negative attribution in addition to positive attribution, unlike the CAM family.
\newline

\noindent \textbf{Integrated Gradients.}
Integrated gradients~\cite{sundararajan2017axiomatic} is an ``axiomatic attribution'' map that satisfies two axioms for input feature relevance scoring on a network $f$. The first axiom is {\em sensitivity:} compared to some baseline input $\boldsymbol{x}'$, when input $\boldsymbol{x}$ differs from $\boldsymbol{x}'$ along feature $x_i$ and $f(\boldsymbol{x}) \neq f(\boldsymbol{x}')$, then $x_i$ should have a non-zero relevance score. The second axiom is {\em implementation invariance}:
for two networks $f_1$ and $f_2$ whose outputs are equal for all possible inputs, the relevance score for every input feature $x_i$ should be identical over $f_1$ and $f_2$. The break of the second axiom may potentially result in the sensitivity of relevance scores on irrelevant aspects of a model.

Given a deep network $f$ whose codomain is $[0,1]$, an input $\boldsymbol{x}$, and a baseline 
input $\boldsymbol{x}'$, the relevance of feature $x_i$ of input $\boldsymbol{x}$ over $f$ is taken as the integral of the gradients of $f$ along the straight line path from $\boldsymbol{x}'$ to $\boldsymbol{x}$: 

\begin{equation}
    IG_i(\boldsymbol{x}) = (x_i - x_i')\int_{0}^1 \frac{\partial f(\boldsymbol{x}'+ \alpha(\boldsymbol{x}-\boldsymbol{x}'))}{\partial x_i} d\alpha
\end{equation}

where $\alpha$ is associated with the path from $\boldsymbol{x}'$ to $\boldsymbol{x}$, and is smoothly distributed in range $[0, 1]$. An interpretation of $IG_i$ is the cumulative sensitivity of $f$ to changes in feature $i$ in all inputs on a straight line between $\boldsymbol{x}'$ to $\boldsymbol{x}$ going in direction $i$. Intuitively, $x_i$ should have increasing relevance if gradients are large between a ``neutral'' baseline point $\boldsymbol{x}'$ and $\boldsymbol{x}$ along the direction of $x_i$. IG can be approximated by a Riemann summation of the integral:

\begin{equation}
    IG_i(\boldsymbol{x}) \approxeq (x_i - x_i')\sum_{k=1}^M \frac{\partial F(\boldsymbol{x}' + \frac{k}{M}(\boldsymbol{x}-\boldsymbol{x}'))}{\partial x_i} \frac{1}{M}
\end{equation}

where $M$ is the number of steps in the Riemann approximation of this integral.
In the original paper the authors propose setting $M$ somewhere between 20 and 300 steps.

IG is similar to LRP and DeepLIFT: the practitioner needs to supply a reference (baseline) image, and the algorithm assigns pixel-wise relevance scores. For image input problems, the baseline image is set to a black image, while for text input, the baseline input can be a zero embedding vector. IG targets the sensitivity axiom not considered by gradient-based attribution methods such as \cite{simonyan2013deep,springenberg2014striving} and the implementation invariance axiom not considered by methods like DeepLIFT and LRP. 

\subsubsection{Perturbation-based Methods}
Perturbation-based methods compute input feature relevance by altering or removing the input feature and comparing the difference in network output between the original and altered one. Perturbation methods can compute the marginal relevance of each feature with respect to how a network responds to a particular input. 
\newline

\begin{figure}
    \centering
    \includegraphics[width=0.95\linewidth]{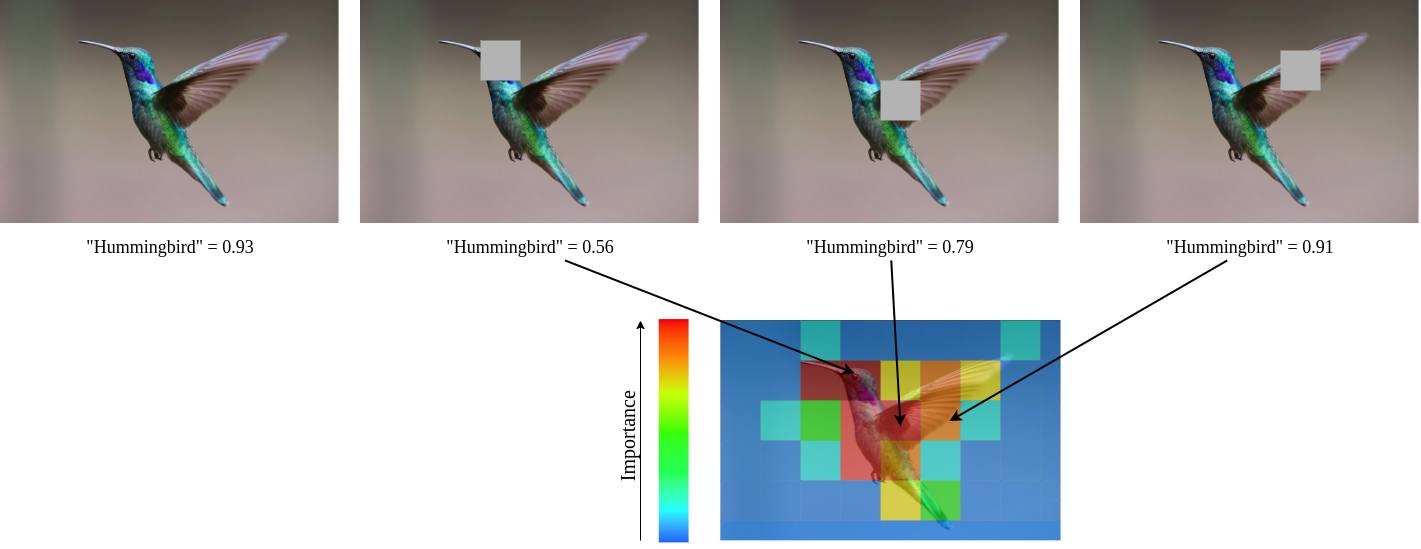}
    \caption{An illustration of how perturbation methods work on images.}
    \label{fig:occlusion_sensitivity}
\end{figure}

\noindent \textbf{Occlusion Sensitivity.}
The approach proposed by \cite{zeiler14}, applicable for spatial data, sweeps a `grey patch' that occludes spatial values (i.e., pixels) over the input and sees how the model prediction varies when the patch covers different regions in the input. The reasoning behind this approach is that the model's performance decreases when the model does not have access to the relevant information. Thus, the more the model performance decreases, the more relevant the occluded region is assumed to be. When a significant portion of the image is swept, the information can be used to create a sensitivity heatmap. In Figure \ref{fig:occlusion_sensitivity} an example is given where a gray patch is swept across an image of a hummingbird. A variant of this method is implemented in~\cite{zhou2014object}, where small gray squares are used to occlude image patches in a dense grid. All other methods in this category work similarly. The methods vary in the information that the patch provides or removes, the size of the patch, and how the patches are sampled. 

The practitioner can use this method to measure how sensitive a model is to a particular part of the removed input. This information can serve as a perfunctory indication of how regions in the input are correlated with model predictions. Since this method does not make use of model internals, it can be used on any DNN. It is worth noting that the higher the desired resolution of heatmaps, the smaller the patch should be, thus the longer it takes to compute the heatmap. Certain features in the input might co-occur, and the joint presence of these features is important. However, occlusion sensitivity is unable to handle this because only one region at a time is occluded.
\newline

\noindent \textbf{Representation Erasure.}
\cite{li2016understanding} is an example of a perturbation-based method for natural language input. To measure the effectiveness of each input word or each dimension of intermediate hidden activations, the method erases the information by deleting a word or setting a dimension to zero and observes the influences on model predictions correspondingly. Reinforcement learning (RL) is adopted to evaluate the influence of multiple words or phrases combined by finding the minimum changes in the text that causes a flipping of a neural network's decision.

Practitioners can use representation erasure to achieve the same goals as occlusion sensitivity: to measure how sensitive a model is to a particular part of the removed input. Even though this method focuses on natural language explanations, it can be modified and applied to other types of problems. Compared to occlusion sensitivity, this method has a couple of benefits. First, representation erasure can handle combinations of erasures. It becomes possible to take into account co-occurring input features. Second, representation erasure is efficient because RL is used to find the minimum change that causes the model's prediction to change. In contrast, occlusion sensitivity requires the practitioner to sweep the entire image in a brute force manner. 
\newline

\noindent \textbf{Meaningful Perturbation.}
\cite{fong2017interpretable} defines an explanation as a meta-predictor, which is a rule that predicts the output of a black box $f$ to certain inputs $x$. For example, the explanation for a classifier that identifies a bird in an image can be defined as 

\begin{equation}
    B(\boldsymbol{x};f) = \left\{\boldsymbol{x} \in \boldsymbol{X}_c \Leftrightarrow f(\boldsymbol{x}) = +1\right\}
\end{equation}

where $f(\boldsymbol{x}) = +1$ means a bird is present and $\boldsymbol{X}_c$ is the set of all images that the DNN predicts a bird exists. Given a specific image $\boldsymbol{x}^0$ and a DNN $f$, the visualization is generated via perturbation to identify sensitive areas of $\boldsymbol{x}^0$ with respect to the output $f(\boldsymbol{x}^0)$ formulated as a local explanation (``local'' to $\boldsymbol{x}^0$) by the author. The author defines three kinds of perturbations to delete information from image, i)~\textit{constant}, replacing region with a constant value ii)~\textit{noise}, adding noise to the region, and iii)~\textit{blur}, blurring the region area, and generating explainable visualization respectively. 

A practitioner can use meaningful perturbation to find regions in the input that the model's output is sensitive to, similar to the previous perturbation-based methods. In contrast to said methods, information is never entirely removed from the image, and the amount of perturbation is kept to a minimum. Because of this, the resulting heatmap is more concentrated around the region of interest, and the number of spurious areas is far less compared to the other methods. From a practitioner's perspective, this makes for more easily interpretable heatmaps. 
\newline

\noindent \textbf{Prediction Difference Analysis.}
\cite{zintgraf2017visualizing} proposes a rigorous approach to delete information from an input and measure its influence accordingly. The method is based on~\cite{robnik2008explaining}, which evaluates the effect of feature $x_i$ with respect to class $c$ by calculating the prediction difference between $p(c \mid \boldsymbol{x}_{-i})$ and $p(c \mid \boldsymbol{x})$ using the marginal probability

\begin{equation}
    p(c \mid \boldsymbol{x}_{-i}) = \sum_{x_i} p(x_i \mid \boldsymbol{x}_{-i})p(c \mid \boldsymbol{x}_{-i}, x_i)
\end{equation}

where $\boldsymbol{x}$ denotes all input features, $\boldsymbol{x}_{-i}$ denotes all features except $x_i$, and the sum iterates over all possible values of $x_i$. The prediction difference, also called {\em relevance value} in the paper, is then calculated by

\begin{equation}
    \textrm{Diff}_i(c \mid \boldsymbol{x}) = \log_2(\texttt{odds}(c\mid \boldsymbol{x})) - \log_2(\texttt{odds}(c \mid \boldsymbol{x}_{-i}))  
\end{equation}

where $\texttt{odds}(c \mid \boldsymbol{x}) = \frac{p(c \mid \boldsymbol{x})}{1 - p(c \mid \boldsymbol{x})}$. The magnitude of $\textrm{Diff}_i(c\mid \boldsymbol{x})$ measures the importance of feature $x_i$.  $\textrm{Diff}_i(c\mid \boldsymbol{x})$ measures the influence direction of feature $x_i$, where a positive value means {\em for} decision $c$ and a negative value means {\em against} decision $c$. Compared to~\cite{robnik2008explaining}, Zintgraf~\etal~improves prediction difference analysis in three ways: by~i) sampling patches instead of pixels given the high pixel dependency nature of images;~ii) removing patches instead of individual pixels to measure the prediction influence given the robustness nature of neutral networks on individual pixels;
~iii) adapting the method to measure the effect of intermediate layers by changing the activations of a given intermediate layer and evaluating the influence on down-streaming layers. The heatmaps produced with this method contain both evidence for and against the predicted class. 

The practitioner can use this method to gain insight into which regions in the input the model is sensitive to and how perturbations in various model regions affect the output. 
Compared to the other perturbation-based methods, prediction difference analysis applies a conditional sampling algorithm to determine which regions are perturbed with Gaussian noise. From a practical perspective, this leads to heatmaps, indicating both positive and negative regions of interest for a specific class. 

\subsection{Model Distillation} 
\label{subsection:model_distillation}

\begin{table}[t]
    \centering
    \resizebox{\textwidth}{!}{
    \begin{tabular}{p{5cm}p{6cm}p{8cm}}
    \specialrule{.2em}{.1em}{.1em}
    \textbf{Model Distillation} & \textbf{Comments} & \textbf{References} \\ 
    \specialrule{.1em}{.1em}{.1em}
    Local Approximation & Learns a simple model whose input/output behavior mimics that of a DNN for a small subset of input data. & \cite{ribeiro2016should,ribeiro2016model,ribeiro2016nothing,ribeiro2018anchors,elenberg2017streaming} \\ \hline
    
    Model  Translation & Train  an  alternative  smaller  model  that  mimics  the input/output behavior of a DNN. & \cite{frosst2017distilling,tan2018learning,zhang2019interpreting,hou2020learning,zhang2017growing,zhang2017interpreting,harradon2018causal,murdoch2017automatic} \\
    
    \specialrule{.2em}{.1em}{.1em}
    \end{tabular}
    }
    \caption{Model distillation.}
    \label{table:model_distillation}
\end{table}

\begin{figure}
    \centering
    \includegraphics[width=0.9\linewidth]{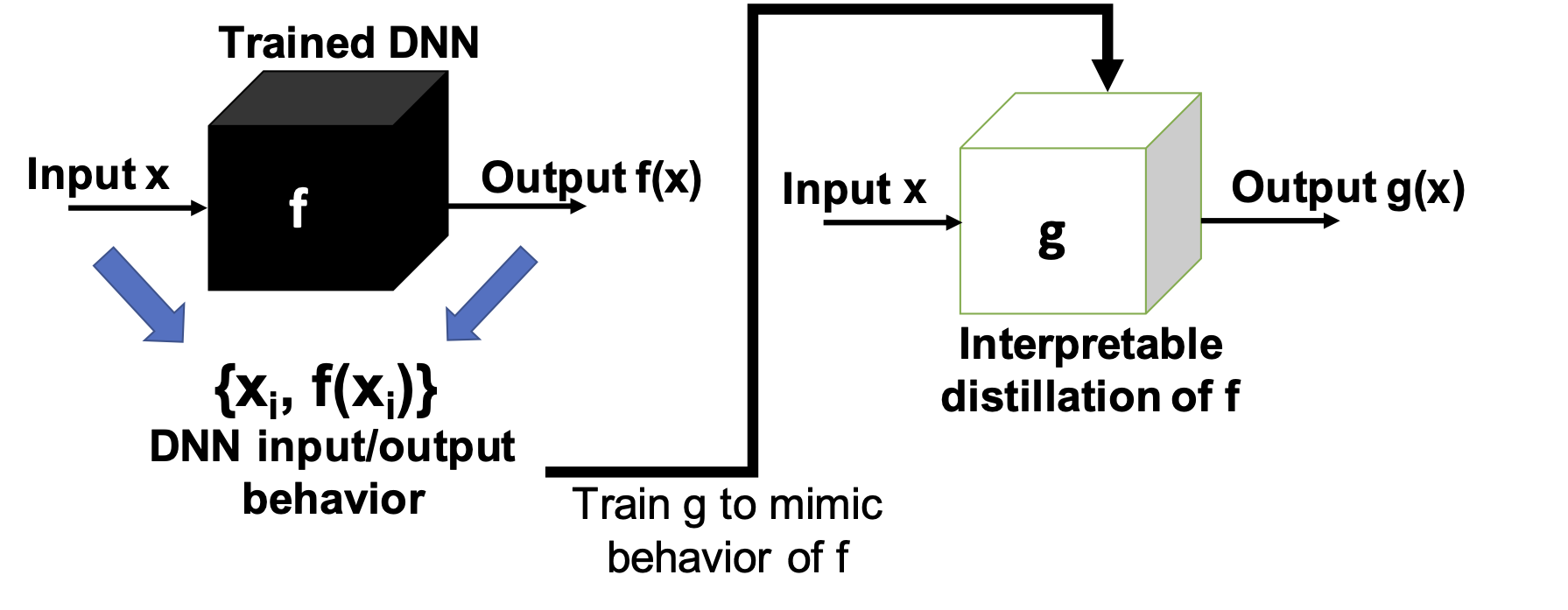}
    \caption{High level view of model distillation. The behavior of a 
    trained deep learning model $f$ used as training data for an 
    explainable model $g$. }
    \label{fig:distillation}
\end{figure}

\textit{Model distillation} refers to a class of post-training explanation methods where the knowledge encoded within a trained DNN is {\em distilled} into a representation amenable for explanation by a user. This representation can take the form of more interpretable machine learning methods, e.g., decision trees. In this setting, as illustrated in Figure~\ref{fig:distillation}, an inherently transparent or white box machine learning model $g$ is trained to mimic the input/output behavior of a trained opaque deep neural network $f$ so that $g(\boldsymbol{x}) \approx f(\boldsymbol{x})$. Subsequent explanation of how $g$ maps inputs to outputs may serve as a surrogate explanation of $f$'s mapping. Note that~\cite{hinton2015distilling} outlines a method, with the same name, that implements a specific form of model distillation, namely distilling the knowledge learned by an ensemble of DNNs into a single DNN. In fact,
an entire class of explainable deep learning techniques has emerged based on the notion of model distillation. 

A distilled model in general learns to imitate the actions or qualities of an opaque DNN over the same data. It is a myth that the distilled form of a DNN necessarily underperforms compared to the original DNN~\cite{liu2018improving}. Conceptually,  this may be because a distilled model has access to information from the trained DNN, including the input features it found to be most discriminatory and feature or output correlations relevant for classification. The distilled model can use this information directly during training, thus reducing the needed capacity of the distilled model. Since the distilled model still takes the original data as input, it may be further useful as a transparent view of how input features become related to the actions of the DNN. Interpreting the distilled model will not give deep insights into the internal representation of the data a DNN learns, or say anything about the DNN's learning process, but can at least provide insight into the features, correlations, and relational rules that explain how the DNN operates. 
Put another way, the explanation of a distilled model can be seen as a hypothesis as to why a DNN has assigned some class label to an input. 

Sometimes distillation methods can appear very similar to occlusion methods because distillation methods can also use occlusion in their algorithms. However, they differ in the following key aspect: While occlusion methods explicitly aim to make heatmaps, distillation methods aim to capture (local) model behavior by linking the information learned by occlusion with a more general representation. Moreover, compared to visualization methods, a much larger space of explanation forms become available to the practitioner that requires specialized knowledge about both the application domain as well as the interpretable model that will be distilled to.

We organize model distillation techniques into two categories: 

\begin{itemize}
    \item {\bf Local Approximation.} 
    A local approximation method learns a simple model whose input/output behavior mimics that of a DNN for a small subset of the input data. 
    This method is motivated by the idea that the model a DNN uses to discriminate within a local area of the data manifold is simpler than the discriminatory model over the entire surface. 
    Given a sufficiently high local density of input data to approximate the local manifold with piecewise linear functions, the DNN's behavior in this local area may be distilled into a set of explainable linear discriminators.
    \item {\bf Model Translation.} 
    Model translations train an alternative smaller model that mimics the total. input/output behavior of a DNN. 
    They contrast local approximation methods in replicating the behavior of a DNN across an entire dataset rather than small subsets. 
    The smaller models may be directly explainable, may be smaller and easier to deploy, or could be further analyzed to gain insights into the causes of the input/output behavior that the translated model replicates. 
\end{itemize}

\subsubsection{Local Approximation}

A local approximation method learns a distilled model that mimics DNN decisions on inputs within a small subset of the input examples.
Local approximations are made for data subsets where feature values are very similar, so that a simple and explainable model can make decisions within a small area of the data manifold. 
While the inability to explain {\em every} decision of a DNN may seem unappealing, it is often the case that an analyst or practitioner is most interested in interpreting DNN actions under a particular subspace (for example, the space of gene data related to a particular cancer or the space of employee performance indicators associated with those fired for poor performance). 

The idea of applying local approximations may have originated from~\cite{baehrens10}. 
These researchers present the notion of an ``explainability vector'', defined by the derivative of the conditional probability a datum is of a class given some evidence $x_0$ by a Bayes classifier. 
The direction and magnitude of the derivatives at various points $x_0$ along the data space define a vector field that characterizes flow away from a corresponding class. 
The work imitates an opaque classifier in a local area by learning a classifier that has the same form as a Bayes estimator for which the explanation vectors can be estimated.
\newline

\begin{figure}
    \centering
    \includegraphics[width=0.95\linewidth]{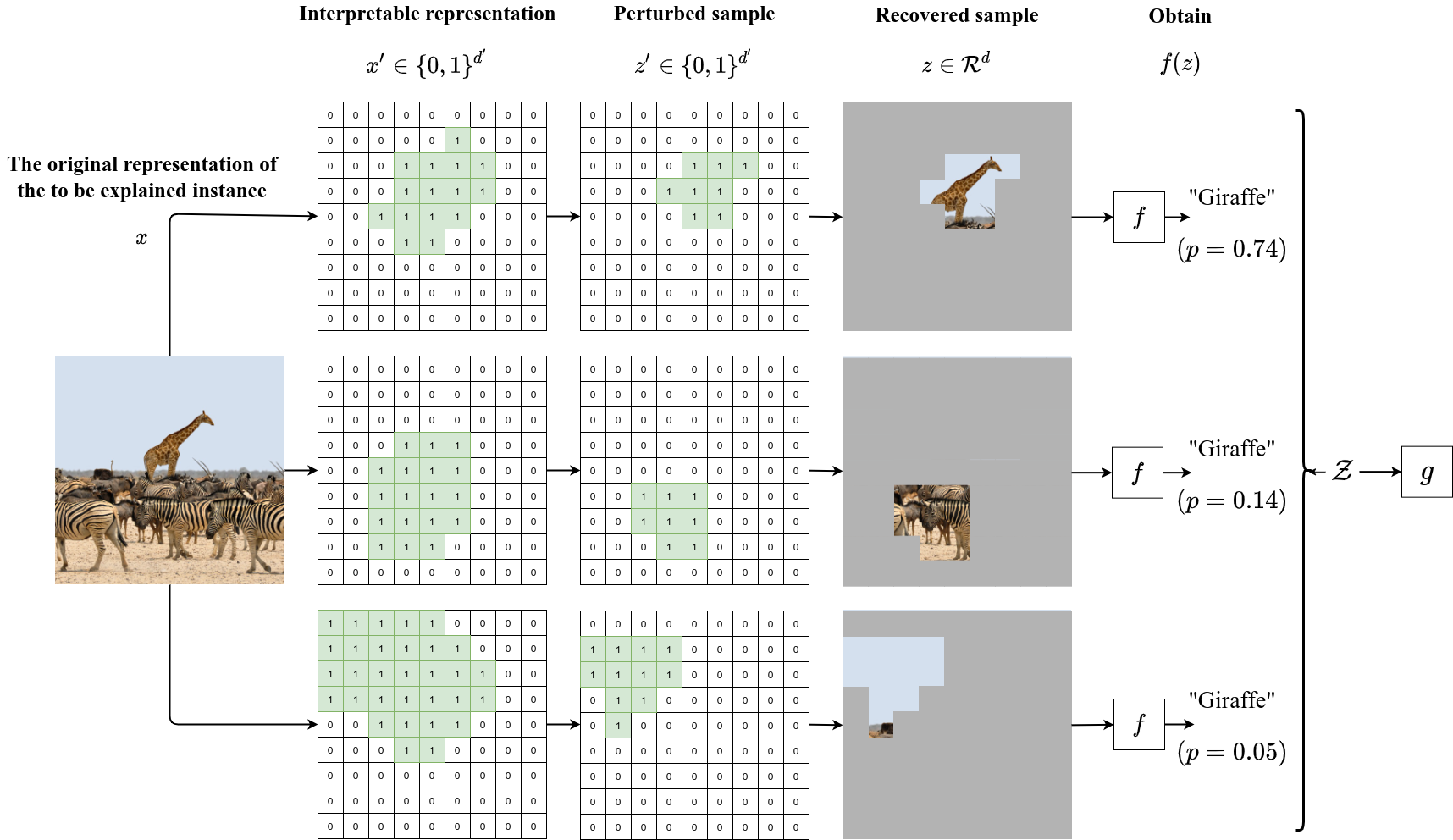}
    \caption{An image example explaining the process behind LIME. The interpretable representation $\boldsymbol{x}'$ is a binary array indicating if a pixel belongs to a pattern or not by respectively assigning a 1 or a 0 to each pixel. $\boldsymbol{z}'$ contains a subset of the nonzero elements of $\boldsymbol{x}'$. $\boldsymbol{z}$ is a perturbed version of $\boldsymbol{x}$ where $\boldsymbol{z}'$ indicates the visible pixels or input features. This process of sampling subsets of the original input is repeated until the desired local perturbed dataset $\mathcal{Z}$ around $\boldsymbol{x}$ is collected. Finally $\mathcal{Z}$ is used to train interpretable model $g$.}
    \label{fig:lime}
\end{figure}

\noindent \textbf{Local Interpretable Model-Agnostic Explanations (LIME).}
Perhaps the most popular local approximation method is \textbf{LIME} developed by~\cite{ribeiro2016should}. Figure \ref{fig:lime} visually outlines the LIME process. From a {\em global}, black-box model $f$ and an input of interest $\boldsymbol{x} \in \mathbb{R}^d$, LIME defines an interpretable model $g$ from a class of {\em inherently} interpretable models $g \in G$ with different domain $\mathbb{R}^{d'}$ that approximates $f$ well in the local area around $\boldsymbol{x}$. Examples of models in $G$ may be decision trees or regression models whose weights explain the relevance of an input feature to a decision. Note that the domain of $g$ is different from that of $f$. The model $g$ operates over an interpretable representation $\boldsymbol{x}'$ of the input data presented to the unexplainable model $f$, which could for example be a binary vector denoting the presence or absence of words in text input, or a binary vector denoting if a certain pixel or color pattern exists in an image input. In Figure \ref{fig:lime} examples of the interpretable representation $\boldsymbol{x}'$ can be seen. In this case $\boldsymbol{x}'$ is a binary array indicating if a pixel belongs to a pattern or not by respectively assigning a 1 or a 0 to each pixel location. Noting that $g$ could be a decision tree with very high depth, or a regression model with many co-variate weights, an interpretable model that is overly complex may still not be useful or usable to a human. Thus, LIME also defines a measure of complexity $\Omega(g)$ on $g$. $\Omega(g)$ could measure the depth of a decision tree or the number of higher order terms in a regression model, for example, or it could be coded as if to check that a hard constraint is satisfied (e.g., $\Omega(g) = \infty$ if $g$ is a tree and its depth exceeds some threshold). Let $\Pi_{\boldsymbol{x}}(\boldsymbol{z})$ be a similarity kernel between perturbed data point $\boldsymbol{z}$ and a original data point $\boldsymbol{x} \in \mathbb{R}^{d}$ and a loss $\mathcal{L}(f,g,\Pi_{\boldsymbol{x}})$ defined to measure how poorly $g$ approximates $f$ on data in the area $\Pi_{\boldsymbol{x}}$ around the data point $\boldsymbol{x}$. To interpret $f(\boldsymbol{x})$, LIME identifies the model $g$ satisfying: 

\begin{equation}
    \arg\min_{g} \left\{ \mathcal{L}(f,g,\Pi_{\boldsymbol{x}}) + \Omega(g) \right\}
    \label{eq:lime}
\end{equation}

where $\Omega(g)$ serves as a type of complexity regularization, or as a guarantee that the returned model will not be too complex when $\Omega(g)$ codes a hard constraint. So that LIME remains model agnostic, $\mathcal{L}$ is approximated by uniform sampling over the non-empty space of $\mathbb{R}^{d'}$. For each sampled data point $\boldsymbol{z}' \in \mathbb{R}^{d'}$, LIME recovers the $\boldsymbol{x} \in \mathbb{R}^d$ corresponding to $\boldsymbol{z}'$, computes $f(\boldsymbol{z})$, and compares this to $g(\boldsymbol{z}')$ using $\mathcal{L}$. To make sure that the $g$ minimizing Equation~\ref{eq:lime} fits well in the area local to the original reference point $\boldsymbol{x}$, the comparison of $f(\boldsymbol{z})$ to $g(\boldsymbol{z}')$ in $\mathcal{L}$ is weighted by $\Pi_{\boldsymbol{x}}(\boldsymbol{z})$ so that samples farther from $\boldsymbol{x}$ contribute less to the loss. The sampling process is repeated until the satisfactory dataset of locally perturbed samples $\mathcal{Z} = \{ \boldsymbol{z}', f(\boldsymbol{z}), \Pi_{\boldsymbol{x}}(\boldsymbol{z}) \} $ is obtained to train interpretable model $g$ on.

LIME gained popularity because it was one of the earlier methods that combined a model-agnostic framework with local model explanations. The strength of LIME lies in the validation of the method with non-expert human practitioners. The original paper describes multiple experiments where non-expert users were asked to use the LIME explanations in different tasks. From an algorithmic perspective, LIME makes use of occlusion to find regions in the input that the model is sensitive to, similar to the perturbation-based methods discussed earlier. However, LIME aims to generalize the explanations in a local region around a reference image by learning an interpretable model. In practice, this means that the practitioner only needs one local LIME model for each set of similar inputs. In contrast to perturbation-based methods, where for each image, a new heatmap is computed, once the practitioner has a local LIME model, assuming the input domain does not drastically change, the practitioner does not need to retrain a new local LIME model. This can be particularly useful when the data distribution has low variance.
\newline

We mention LIME in detail because other popular local approximation models~\cite{ribeiro2016model,ribeiro2016nothing,ribeiro2018anchors,elenberg2017streaming} follow LIME’s template and make their extensions or revisions. One drawback of LIME, which uses a linear combination of input features to provide local explanations, is the precision and coverage of such explanations are not guaranteed. Since the explanations are generated in a locally linear fashion, for an unseen instance, which might lie outside of the region where a linear combination of input features could represent, it is unclear if an explanation generated linearly and locally still applies. To address this issue, {\em anchor} methods~\cite{ribeiro2016nothing,ribeiro2018anchors} extend LIME to produce local explanations based on if-then rules, such that the explanations are locally anchored upon limited yet sufficiently stable input features for the given instance and the changes to the rest of input features won’t make an influence. Another drawback of LIME is, the interpretable linear model is trained based on a large set of randomly perturbed instances and the class label of each perturbed instance is assigned by inevitably calling the complex opaque model, which is computationally costly. To reduce the time complexity, \cite{elenberg2017streaming} introduces STREAK, which is similar to LIME but limits the time of calling complex models, and thus runs much faster. Instead of randomly generating instances and training an interpretable linear model, STREAK directly selects critical input components (for example, superpixels of images) by greedily solving a combinatorial maximization problem. Taking the image classification task as an example, an input image which is predicted as a class by the opaque model, is first segmented into superpixels via image segmentation algorithm~\cite{achanta2012slic}. In every greedy step, a new superpixel is added to the superpixels set if by containing it in the image will maximize the probability of the opaque model on predicting the given class. The set of superpixels indicating the most important image regions of the given input image for the opaque model to make its decision. Despite the technical details, some common characteristics are shared among all aforementioned local approximation methods, 
i) input instance is segmented into semantic meaningful parts for selection, 
ii) function calls of the original opaque model is inevitable 
iii) the explanations and model behavior are explored in a local fashion.
\newline

\noindent \textbf{Shapley Additive Explanations.} More commonly referred to as SHAP~\cite{lundberg2017unified}, this method computes Shapley values~\cite{shapley1953value} for input feature sets. The Shapley value explanations are represented as the coefficients in a linear model. SHAP bears a resemblance to perturbation-based methods because an incomplete (perturbed) input $z$ is given to the model. The effects of the perturbation are measured, and a score is assigned based on the feature contribution amount. Only in this case, the contribution of \textit{adding} a feature is measured and as opposed to measuring the \textit{removal} of a feature as is the case in perturbation-based methods. The paper presents a framework grounded in game theory that provides a result guaranteeing a unique solution for additive feature attribution methods. According to the paper, various other explanation methods can be considered additive feature attribution methods. 

\begin{figure}
    \centering
    \includegraphics[width=0.95\linewidth]{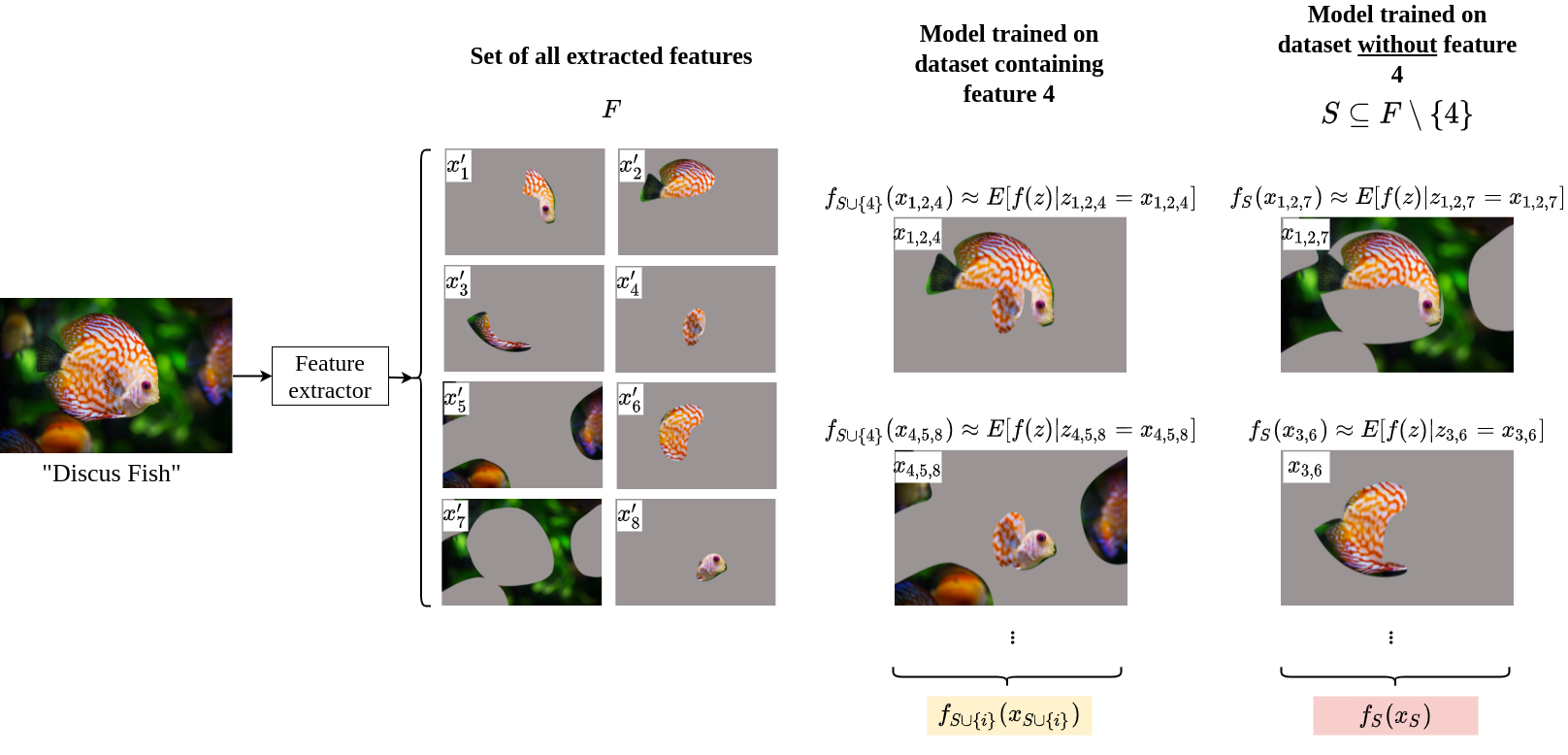}
    \caption{A visualization of the SHAP algorithm. In this example we have $M=8$ simplified input features and the contribution of feature $x'_4$ is being measured.}
    \label{fig:SHAP}
\end{figure}

In essence, each feature is viewed as a member of a group. The method calculates how much each member contributes to the group. At the heart of the SHAP lies a function that assigns contribution values to each input feature. These are called the Shapley values $\phi_i$:

\begin{equation}
    \phi_i = \sum_{S \subseteq F \setminus \{i\}}  \frac{|S|!(|F|-|S|-1)!}{|F|!} [f_{S \cup \{i\}}(x_{S \cup \{i\}
})-f_S(x_S)]
\end{equation}

where $F$ is the set of all features, $S \subseteq F$, $f_{S \cup \{i\}}(x_{S \cup \{i\}})$ is a model trained with a subset of features that does \emph{not} include feature $x'_i$ and $f_S(x_S)$ is a model trained with a subset of features that does include feature $x'_i$. \noindent In practice this is implemented as:

\begin{equation}
    \phi_i(f, x) = \sum_{z' \subseteq x'} \frac{ |z'|!(M-|z'|-1)! }{M!} [ f_x(z') - f_x(z' \setminus i) ]
\end{equation}

where $z' \in \{ 0, 1\}^M$ is a simplified representation of the perturbed sample $z$ indicating the presence of input features $x_i'$ and $M$ is the number of simplified input features. $\phi_i$ is calculated by sampling various combinations of features and measuring the change in prediction. In the end, a linear model $g$ is fitted to the features and their effects:

\begin{equation}
    g(z') = \phi_0 + \sum^M_{i=1} \phi_i z'_i
    \label{eq:lin}
\end{equation}
 An example of this process is illustrated in Figure \ref{fig:SHAP}. 

There exist different implementations of SHAP. KernelSHAP repeatedly samples features from the input, replacing a subset of the values by random values present in the data. This perturbed input is fed into the model, and the prediction is recorded. Each sampled feature set is assigned a weight using the SHAP kernel. After the sampling is done, a linear model is fitted. The Shapley values are extracted as the coefficients from the linear model. KernelSHAP samples from the marginal distribution and assumes that features are independent from each other. However, this is often not the case with real-world data such as naturalistic images. TreeSHAP was introduced as a faster alternative to KernelSHAP~\cite{lundberg2018consistent}. Note that TreeSHAP is applicable only to tree-based ML models. The critical difference and why we mention TreeSHAP is that it samples from the conditional distribution instead of the marginal distribution, i.e., it does not assume that features are independent. Features that on themselves are not relevant for prediction can get a non-zero value assigned because they can now be correlated to other features relevant for model prediction. This is problematic because it violates the Shapley axiom stating that features that do not contribute to the prediction should have a value of zero. In practice, it means that the explanations produced by TreeSHAP are unreliable. \cite{heskes2020causal} proposes a causal variation on SHAP called causal Shapley values. Causal Shapley values give a causal interpretation to Shapley values, allowing for the differentiation between direct and indirect feature contributions. 

Even though SHAP is considered a local approximation method, we can run it multiple times to obtain global explanations. This is where SHAP becomes a compelling method since the global explanations are faithful to the local ones. Assuming sufficient Shapley values are calculated, SHAP becomes an explanation model in itself that can explain any instance. Strong examples are given by \cite{molnar2020interpretable}. 

As mentioned before, SHAP uses occlusion to find regions in the input that the model's output is sensitive to, similar to LIME. The other similarity that they share is that the explanation is an interpretable model. In the case of LIME, the choice of model is left to the practitioner, while with SHAP, the interpretable model is always a linear model as defined in Equation \ref{eq:lin}. SHAP's main strength comes from the way it defines an explanation as additive feature attribution grounded in a game-theoretical perspective. Viewed through this lens, various explanation methods fit into this framework, and the relationships between methods become apparent. Like LIME, SHAP was also validated with real human practitioners, and the results show that SHAP explanations are better aligned with human intuition compared to several other methods, including LIME. 

\subsubsection{Model Translation}
Compared to local approximation methods, model translation replicates the behavior of a DNN across an \textit{entire} dataset rather than small subsets, through a smaller model that is easier for explanation. The smaller model could be easier to deploy~\cite{hinton2015distilling}, faster to converge~\cite{yim2017gift}, or even be easily explainable, such as a decision tree~\cite{frosst2017distilling,bastani2017interpreting,tan2018learning}, Finite State Automaton (FSA)~\cite{hou2020learning}, graphs~\cite{zhang2017growing,zhang2017interpreting}, or causal and rule-based classifier~\cite{harradon2018causal,murdoch2017automatic}. We highlight the diversity of model types DNNs have been distilled into below.
\newline

\noindent \textbf{Distillation to Decision Trees.}
Recent work has been inspired by the idea of tree-based methods for DNNs. \cite{frosst2017distilling} proposes ``soft decision trees'' which use stochastic gradient descent for training based on the predictions and learned filters of a given neural network. The performance of the soft decision trees is better than normal trees trained directly on the given dataset, but is worse compared to the given pre-trained neural networks. Another recent work is proposed by \cite{tan2018learning} which generates global additive explanations for fully connected neural networks trained on tabular data through model distillation. Global additive explanations~\cite{sobol2001global,hooker2004discovering,hoos2014efficient} have been leveraged to study complex models, including analyzing how model parameters influence model performance and decomposing black box models into lower-dimensional components. In this work, the global additive explanations are constructed by following previous work~\cite{hooker2007generalized} to decompose the black-box model into an additive model such as spline or bagged tree. They follow~\cite{craven1996extracting} to train the additive explainable model. \cite{zhang2019interpreting} trains a decision tree to depict the reasoning logic of a prep-trained DNN with respect to given model predictions. The authors first mine semantic patterns, such as objects, parts, and ``decision modes'' as fundamental blocks to build the decision tree. The tree is then trained to quantitatively explain which fundamental components are used for a prediction and the percentage of contribution respectively. The decision tree is organized in a hierarchical coarse-to-fine way, thus nodes close to the tree top correspond to common modes shared by multiple examples, while nodes at the bottom represent fine-grained modes with respect to specific examples.

From a practitioner's perspective, distilling into decision trees has some advantages, arguably the most important one being that, compared to potentially subjective heatmaps, trees produce objective rules. Furthermore, formulating explanations as rules has the benefit that other algorithms can directly interpret the rules. The practitioner can then further automate the processing of the resulting explanations in their decision systems.
\newline

\noindent \textbf{Distillation to Finite State Automata.}
\cite{hou2020learning} introduces a new distillation of RNNs to explainable Finite State Automata (FSA). It is worth noting that this method is very specialized, only applicable to RNNs that perform binary classifications. An FSA consists of finite states and transitions between the states, and the transition from one state to another is a result of external input influence. FSA is formally defined as a 5-tuple $(\mathbb{E}, \mathbb{S}, s_0, \delta, \mathbb{F})$, where $\mathbb{E}$ is a finite non-empty set of elements existing in input sequences, $\mathbb{S}$ is a finite non-empty set of states, $s_0 \in \mathbb{S}$ is an initial state, $\delta : \mathbb{S} \times \mathbb{E} \rightarrow \mathbb{S}$ defines the state transmission function, and $F \subseteq S$ is the set of final states. The transition process of FSA is similar to RNNs in the sense that both methods accept items from some sequence one by one and transit between (hidden) states accordingly. The idea to distillate an RNN to FSA is based on the fact that the hidden states of an RNN tend to form clusters, which can be leveraged to build FSA. Two clustering methods, \textit{k-means++} and \textit{k-means-x} are adopted to cluster the hidden states of RNN towards constructing the explainable FSA model. The authors follow the structure learning technique and translate an RNN into an FSA, which is easier to interpret in two aspects, i) FSA can be simulated by humans; 
ii) the transitions between states in FSA have real physical meanings. Such a model translation helps to understand the inner mechanism of the given RNN model. 

This is one of the few methods that try to understand the inner states of RNNs. The main benefit of distilling into an FSA is that an FSA can be represented graphically and in an objective manner. Similar to rules produced by a decision tree, this leads to objective explanations.
\newline

\noindent {\bf Distillation into Graphs.} 
Both \cite{zhang2017growing} and \cite{zhang2017interpreting} build an object parts graph for a pre-trained CNN to provide model explanations. Similar to~\cite{zhang2019interpreting}, the authors first extract semantic patterns in the input and then gradually construct the graph for an explanation. Each node in the graph represents a part pattern, while each edge represents co-activation or spatial adjacent between part patterns. The explanatory graph explains the knowledge hierarchy inside the model, which can depict which nodes/part patterns are activated and the location of the parts in the corresponding feature maps.

The benefit of distilling into graphs is that graphs can capture and make transparent relational data. For this method specifically, the information contained in the graph can be represented in several ways, e.g., as intuitive heatmaps that show which semantically interpretable part of the image was relevant to the prediction of the original model, or a graph that displays the connections between features in the input and how these connections relate to the model prediction.
\newline

\noindent \textbf{Distillation into Causal and Rule-based Models.}
We also note work on distilling a DNN into symbolic rules and causal models. 
\cite{harradon2018causal} constructs causal models based on concepts in a DNN. The semantics are defined over an arbitrary set of ``concepts'', that could range from recognition of groups of neuron activations up to labeled semantic concepts. To construct the causal model, concepts of intermediate representations are extracted via an autoencoder. Based on the extracted concepts, a graphical Bayesian causal model is constructed to build association for the models' inputs to concepts, and concepts to outputs. The causal model is finally leveraged to identify the input features of significant causal relevance with respect to a given classification result.

The practitioner can use this method to relate the model prediction to the model's learned concepts. So far, the other methods that do this as well were activation minimization and deconvolution. It is difficult to say whether one method is better than the other because the approaches vary significantly: activation minimization generates an image that maximally activates a target region in the network, sometimes producing strange images. Deconvolution uses translated convolutions to map the target activations back to the input, resulting in patterns representing concepts in the input. The difference between these two methods and \cite{harradon2018causal} is that the latter visualizes concepts as heatmaps overlaid on the input, rather than generating a potentially uninterpretable image. However, this approach can sometimes lead to coarse explanations, especially if a concept is a relatively small part of the input. Another drawback of this method is that the practitioner needs to provide semantic concept labels, which is not necessary with the two previously mentioned methods.

In another example, \cite{murdoch2017automatic} leverages a simple rule-based classifier to mimic the performance of an LSTM model. This study runs experiments on two natural language processing tasks, sentiment analysis, and question answering. The rule-based model is constructed via the following steps:
i) decompose the outputs of an LSTM model, and generate important scores for each word; 
ii) based on the word level important score, important simple phrases are selected according to which jointly have high important scores;
iii) The extracted phrase patterns are then used in the rule-based classifier, approximating the output of the LSTM model.
Similar to using decision trees, rules have the advantage of being objectively interpretable and can be further processed as part of a decision system. 

\subsection{Intrinsic Methods} 
\label{subsection:intrinsic_methods}

\begin{table}[t]
    \centering
    \resizebox{\textwidth}{!}{
    \begin{tabular}{p{6cm}p{6cm}p{8cm}}
    \specialrule{.2em}{.1em}{.1em}
    \textbf{Intrinsic Methods} & \textbf{Comments} & \textbf{References} \\ 
    \specialrule{.1em}{.1em}{.1em}
    
    Attention Mechanisms & Leverage attention mechanisms to learn conditional distribution over given input units, composing a weighted contextual vector for downstream processing. The attention visualization reveals inherent explainability. & \cite{bahdanau2014neural,luong2015effective,vaswani2017attention,wang2016attention,letarte2018importance,he2018effective,devlin2019bert,vinyals2015show,xu2015show,antol2015vqa,goyal2017making,teney2018tips,mascharka2018transparency,anderson2018bottom,xie2019visual,park2016attentive} \\ \hline
    
    Joint Training & Add additional explanation ``task'' to the original model task, and \textit{jointly train} the explanation task along with the original task. & \cite{zellers2019recognition,liu2019towards,park2018multimodal,kim2018textual,hendricks2016generating,camburu2018snli,hind2019ted,melis2018towards,iyer2018transparency,lei2016rationalizing,dong2017improving,li2018deep,chen2019looks} \\ 
    
    \specialrule{.2em}{.1em}{.1em}
    \end{tabular}
    }
    \caption{Intrinsic methods.}
    \label{table:intrinsic_methods}
\end{table}

\begin{figure}
    \centering
    \includegraphics[width=0.95\linewidth]{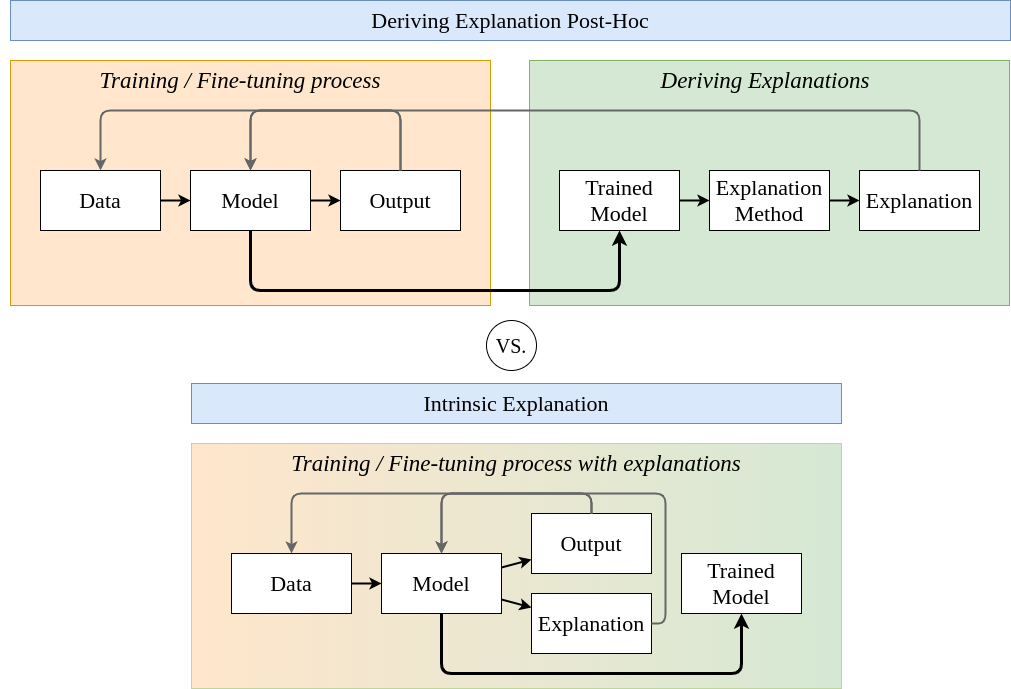}
    \caption{A process comparison between deriving an explanation post-hoc vs. a model where the explanations are intrinsic. The goal is to produce a trained, explainable model. When extracting explanations post-hoc, the process consists of separate training and explanation stages. The explanations can potentially guide the original model for performance improvement (grey arrows going backward). For intrinsic explanation, generating explanations is integrated into the training process. It is worth noting that using intrinsic explanations could be more time-consuming since the model needs to learn a more complex problem.}
    \label{fig:posthocvsintrinsic}
\end{figure}

Ideally, we would like to have models that provide explanations for their decisions as part of the model output, or that the explanation can easily be derived from the model architecture. In other words, explanations should be \textit{intrinsic} to the process of designing model architectures and during training. The ability for a network to intrinsically express an explanation may be more desirable compared to post-hoc methods that seek explanations of models that were never designed to be explainable in the first place. This is because an intrinsic model has the capacity to learn not only accurate outputs per input but also outputs expressing an explanation of the network's action that is optimal with respect to some notion of explanatory fidelity. \cite{ras2018explanation} previously defined a category related to this approach as \textit{intrinsic methods} and identified various methods that offer explainable extensions of the model architecture or the training scheme. In this section, we extend the notion of intrinsic explainability with models that actually provide an explanation for their decision even as they are being trained. Figure \ref{fig:posthocvsintrinsic} shows the difference between the process of deriving explanations post-hoc compared to intrinsic explanations. The main difference is that with intrinsic explanations, the explanations are part of the model or part of the model output, giving the practitioner additional information during the model training phase that is not available during the post-hoc derivation of explanation.

Like distillation methods, the practitioner needs to have explicit knowledge about the field of application and the models required to learn explanations intrinsically. Intrinsic methods are arguably the most difficult methods to apply compared to visualization and distillation methods because the practitioner needs additional specialized knowledge and because the training process will involve multiple models that can make optimization more challenging. Additionally, more computing power is likely required to train a larger pipeline of models jointly. 

We observe methods in the literature on intrinsically explainable DNNs to follow two trends: 
(i) they introduce \textit{attention mechanisms} to a DNN, and the attention visualization reveals inherent explainability; 
(ii) they add additional explanation ``task'' to the original model task, and \textit{jointly train} the explanation task along with the original task.
We explain the trends and highlight the representative methods below.

\subsubsection{Attention Mechanisms}
DNNs can be endowed with attention mechanisms that simultaneously preserve or even improve their performance and have explainable outputs expressing their operations. An attention mechanism~\cite{vaswani2017attention,devlin2019bert,teney2018tips,xie2019visual} learns conditional distribution over given input units, composing a weighted contextual vector for downstream processing. The attention weights can be generated in multiple ways, such as by calculating cosine similarity~\cite{graves2014neural}, adding additive model structure, such as several fully connected layers, to explicitly generate attention weights~\cite{bahdanau2014neural}, leveraging the matrix dot-product~\cite{luong2015effective} or scaled dot-product~\cite{vaswani2017attention}, and so on. Attention mechanisms have shown to improve DNN performance for particular types of tasks, including tasks on ordered inputs as seen in natural language processing~\cite{vaswani2017attention,devlin2019bert} and multi-modal fusion such as visual question answering~\cite{anderson2018bottom}. It is worth noting that recently some interesting discussions have been raised on whether or not attention can be counted as an explanation tool~\cite{jain2019attention,wiegreffe2019attention}; this will be discussed in Section \ref{section:evaluation}.
\newline

\begin{figure}
    \centering
    \includegraphics[width=0.95\linewidth]{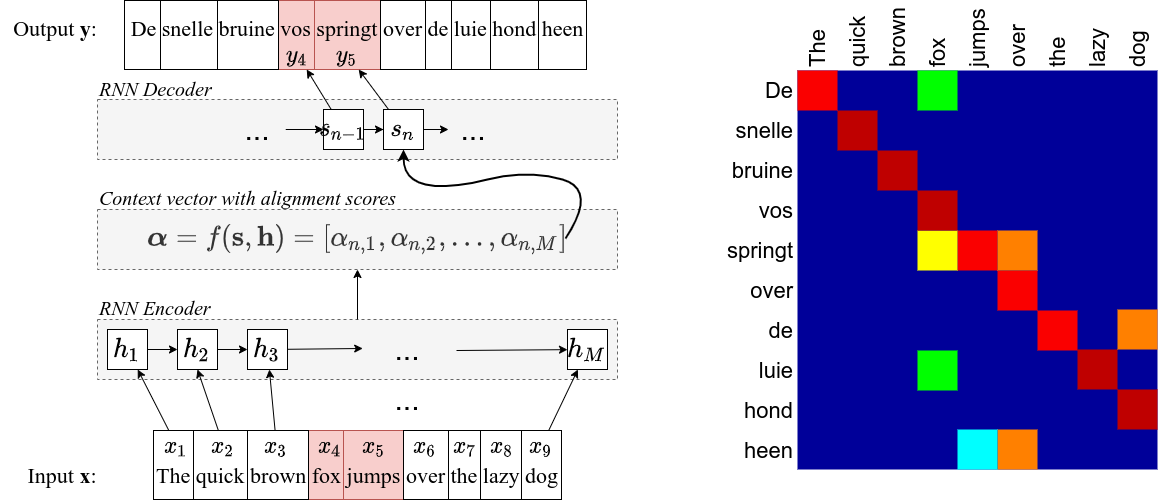}
    \caption{\textbf{Left}: High-level overview of using attention as explanation. The model takes an English sentence as input and outputs a Dutch translation. During the forward pass, the alignment scores $\boldsymbol{\alpha}$ (attention weights) are calculated as part of the training process and can immediately be visualized as a heatmap. The $\boldsymbol{\alpha}$ maps the correlation between the different parts of the input (hidden states $\mathbf{h}$) and output (hidden states $\mathbf{s}$). Alignment score function $f$ determines how $\boldsymbol{\alpha}$ is computed. For a detailed explanation on how attention works, see \cite{bahdanau2014neural}. \textbf{Right}: The attention matrix where each value is the alignment score $\alpha_{n, m}$ between encoder hidden state $h_m$ and decoder hidden state $s_n$.}
    \label{fig:attention}
\end{figure}

\noindent \textbf{Single-Modal Weighting.} 
The output of attention mechanisms during a forward pass can inform a practitioner about how different input features are weighted at different phases of model inference. In pure text processing tasks such as language translation~\cite{bahdanau2014neural,luong2015effective,vaswani2017attention} or sentiment analysis~\cite{wang2016attention,letarte2018importance,he2018effective}, attention mechanism allows the downstream modules, a decoder for language translation or fully connected layers for classification tasks, to concentrate on different words in the input sentence by assigning learned weights to them~\cite{vaswani2017attention,wang2016attention}. To provide straightforward explanations, the attention weights can be visualized as heatmaps, depicting the magnitude and the sign (positive or negative) of each weight value, showing how input elements weighted combined to influence the model latter processing and the final decisions. 

\indent Figure \ref{fig:attention} gives a visual example of how attention can be used for explanation for single-modal weighting. The left-hand side of the figure depicts a basic RNN encoder-decoder architecture where the dense representation of text input is associated with the attention weights. These weights can be plotted as a matrix, and each input's importance as it relates to the output can directly be interpreted. This type of explanation can help the practitioner monitor model predictions during training and give insight into whether the model utilizes undesirable correlations in the dataset. However, the practitioner needs to keep in mind that it will become cumbersome to monitor individual model predictions as the input size increases.
\newline

\noindent \textbf{Multi-Modal Interaction.}
In multi-modal interaction tasks, such as image captioning~\cite{vinyals2015show,xu2015show}, visual question answering~\cite{antol2015vqa,goyal2017making,johnson2017clevr,teney2018tips} or visual entailment~\cite{xie2019visual}, attention mechanisms play an important role in feature alignment and fusion across different feature spaces (for instance, between text and images). 
For example, Park {\em et al.} propose the Pointing and Justification model that uses multiple attention mechanisms to explain the answer of a VQA task with natural language explanations and image region alignments~\cite{park2016attentive}. Xie {\em et al.} use attention mechanisms to recover semantically meaningful areas of an image that correspond to the reason a statement is, is not, or could be entailed by the image's conveyance~\cite{xie2019visual}. \cite{mascharka2018transparency} aims to close the gap between performance and explainability in visual reasoning by introducing a neural module network that explicitly models an attention mechanism in image space. By passing attention masks between modules it becomes explainable by being able to directly visualize the masks. This shows how the attention of the model shifts as it considers the different components of the input.

Multi-modal interaction methods go one step beyond single-modal weighting by combining multiple attention mechanisms with supplementary tasks that increase (i) the model's interpretability and (ii) give the practitioner additional opportunities for creating explanations that cater to the area of application. However, compared to single-modal weighting, multi-modal interaction can be more difficult to apply due to the higher complexity accompanying the increased combination of attention components and multiple tasks. 

\subsubsection{Joint Training}
This type of intrinsic method is to introduce an additional ``task'' besides the original model task, and jointly train the additional task together with the original one. Here we generalize the meaning of a ``task'' by including preprocessing or other steps involved in the model optimization process. The additional task is designed to provide model explanations directly or indirectly. Such additional task can be in the form of 
i) text explanation, which is a task that directly provides explanations in natural language format;
ii) explanation association, which is a step that associates input elements or latent features with human-understandable concepts or objects, or even directly with model explanations; 
iii) model prototype which learns a prototype that has clear semantic meanings as a preprocessing step. Explanations are generated based on the comparison between the model behavior and the prototype.

\begin{figure}
    \centering
    \includegraphics[width=0.95\linewidth]{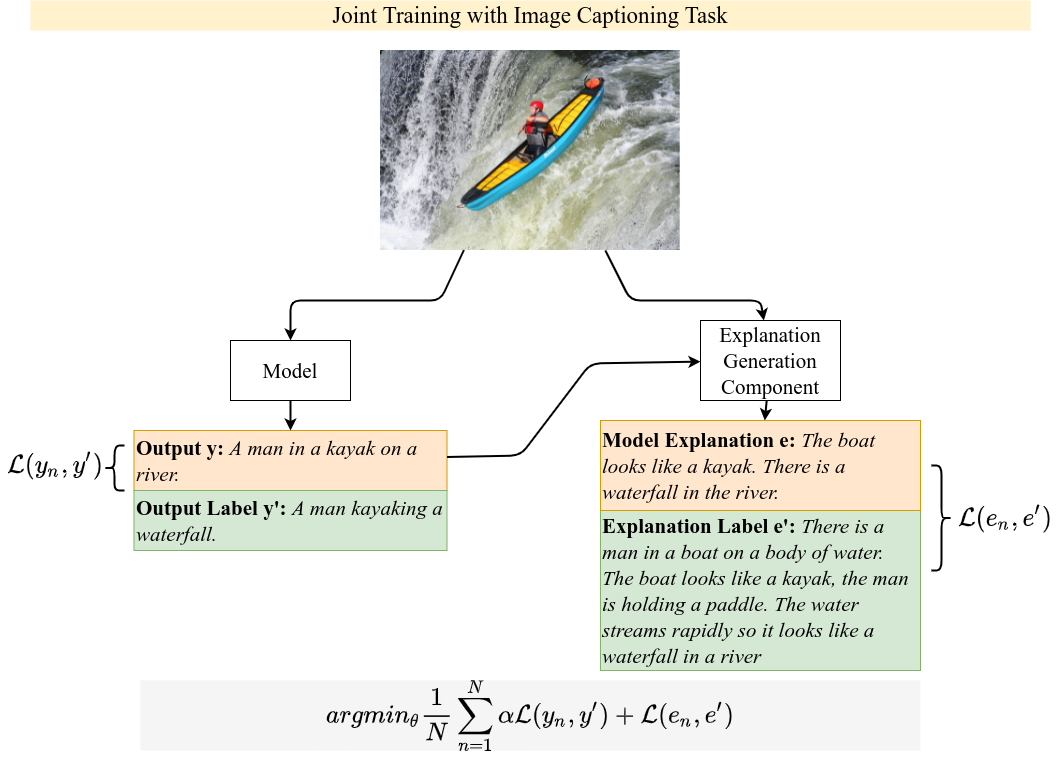}
    \caption{High level overview of how joint training works. The algorithm tries to minimize the average combined weighted loss over the output prediction and the explanation generation, where $\alpha$ denotes the weight.}
    \label{fig:joint}
\end{figure}

In Figure \ref{fig:joint} we see a high-level overview of the joint training with, in this case, an example of an image captioning task. The objective

\begin{equation}
    \arg\min_{\theta}  \frac{1}{N}\sum_{n=1}^N \alpha\mathcal{L}(y_n, y')+\mathcal{L}(e_n, e')
\end{equation}

is a very general form of the function that has to be minimized and is composed of at least two losses: the prediction loss and the explanation component loss. By weighting each loss, a balance can be found between having a model that gives good predictions and a model that gives good explanations. 
\newline

\noindent {\bf Text Explanation.}
A group of recent work~\cite{zellers2019recognition,liu2019towards,park2018multimodal,kim2018textual,hendricks2016generating,camburu2018snli,hind2019ted} achieve the explainable goal via augmenting the original DNN architecture with an explanation generation component and conducting joint training to provide natural language explanations along with the model decisions, similar to what is illustrated in Figure \ref{fig:joint}. 
Such explainable methods are quite straightforward and the explanations that they produce are layman-friendly since the explanations are presented directly using natural language sentences, instead of figures or statistical data that usually require professional knowledge to digest. The explanation could be either generated word by word similar to a sequence generation task~\cite{hendricks2016generating,park2018multimodal,camburu2018snli,kim2018textual,liu2019towards}, or be predicted from multiple candidate choices~\cite{zellers2019recognition}. 

The primary advantage of joint training text explanations is that the practitioner can tailor the explanations to the users' needs while using state-of-the-art models. This way, the practitioner can get the best of both worlds. On the other hand, obtaining the appropriate (labeled) dataset for the explanation generation component is difficult and time-consuming. Furthermore, the practitioner must overcome the additional difficulties that joint training presents when training multiple models jointly instead of training a single model. Finally, it is known that the generated explanations exhibit some inconsistencies~\cite{oana2019make} which undermines the trust in the explanations provided by the model. 

\cite{hendricks2016generating} is an early work that provides text justifications along with its image classification results.
The approach combines image captioning, sampling, and deep reinforcement learning to generate textual explanations. 
The class information is incorporated into the text explanations, which makes this method distinct from normal image captioning models that only consider visual information, via i) include class as an additional input for text generation and ii) adopt a reinforcement learning based loss that encourages generated sentences to include class discriminative information.

\cite{liu2019towards} proposes a Generative Explanation Framework (GEF) for text classifications. The framework is designed to generate fine-grained explanations such as text justifications. During training, both the class labels and fine-grained explanations are provided for supervision, and the overall loss of GEF contains two major parts, classification loss and explanation generation loss. To make the generated explanations class-specific, ``explanation factor'' is designed in the model structure to associate explanations with classifications. The ``explanation factor'' is intuitively based on directly taking the explanations as input for classification and adding constraints on the classification softmax outputs. Specifically, ``explanation factor'' is formulated to minimize the pairwise discrepancy in softmax outputs for different input pairs, i) generated explanations and ground-truth explanations, and ii) generated explanations and original input text. 

Unlike the previously methods which \textit{generate} text explanations, \cite{zellers2019recognition} provides explanations in a \textit{multichoice} fashion. They propose a visual reasoning task named Visual Commonsense Reasoning (VCR), which is to answer text questions based on given visual information (image), and provide reasons (explanations) accordingly. Both the answers and reasons are provided in a multichoice format. Due to the multichoice nature, reasonable explanations should be provided during testing, in contrast to other works which could generate explanations along with model decisions. Thus VCR is more suitable to be applied for prototype model debugging to audit model reasoning process, instead of real-life applications where explanations are usually lacking and remain to be generated.
\newline

\noindent {\bf Explanation Association.}
This type of joint training method associates input elements or latent features with human-understandable concepts or objects, or even directly with model explanations, which helps to provide model explanations intrinsically~\cite{melis2018towards,iyer2018transparency,lei2016rationalizing,dong2017improving}. Such methods usually achieve explanations by adding regularization term~\cite{melis2018towards,lei2016rationalizing,dong2017improving} and/or revising model architecture~\cite{melis2018towards,iyer2018transparency,lei2016rationalizing}. The explanations are provided in the form of 
i) associating input features or latent activations with semantic concepts~\cite{melis2018towards,dong2017improving};
ii) associating model prediction with a set of input elements~\cite{lei2016rationalizing};
iii) associating explanations with object saliency maps in a computer vision task~\cite{iyer2018transparency}.
Regardless of the format of explanations and the technical details, methods belonging to this type commonly share the characteristics of associating hard-to-interpret elements to human-understandable atoms in an intrinsic joint training fashion.

\cite{melis2018towards} proposes an intrinsic method which associates input features with semantically meaningful concepts and regards the coefficient as the importance of such concepts during inference. A regularization based general framework for creating self-explaining neural networks (SENNs) is introduced. Given raw input, the network jointly learns to generate the class prediction and to generate explanations in terms of an input feature-to-concept mapping. The framework is based on the notion that linear regression models are explainable and generalizes the respective model definition to encompass complex classification functions, such as a DNN. A SENN consists of three components: 
i) A ``concept encoder'' that transforms the raw input into a set of explainable concepts. Essentially this encoder can be understood as a function that transforms low-level input into high-level meaningful structure, which predictions and explanations can be built upon. 
ii) An ``input-dependent parametrizer'', which is a procedure to get the coefficient of explainable concepts, learns the relevance of the explainable concepts for the class predictions. The values of the relevance scores quantify the positive or negative contribution of the concept to the prediction. 
iii) Some ``aggregation function'' (e.g. a sum) that combines the output of the concept encoder and the parametrizer to produce a class prediction. 

\cite{iyer2018transparency} introduces Object-sensitive Deep Reinforcement Learning (O-DRL), which is an explanation framework for reinforcement learning tasks that takes videos as input. O-DRL adds a pre-processing step (template matching) to recognize and locate specific objects in the input frame. For each detected object, an extra channel is added to the input frame's RGB channels. Each object channel is a binary map that has the same height and width as the original input frame, 1's encoding for the location of the detected object. The binary maps are later used to generate object saliency maps (as opposed to pixel saliency maps) that indicate the relevance of the object to action generation. It is argued that object saliency maps are more meaningful and explainable than pixel saliency maps since the objects encapsulate a higher-level visual concept.

\cite{lei2016rationalizing} integrates explainability in their neural networks for sentiment analysis by learning rationale extraction during the training phase in an unsupervised manner. Rationale extraction is done by allowing the network to learn to identify a small subset of words that all lead to the same class prediction as the entire text. They achieve this by adding mechanisms that use a combination of a generator and an encoder. The generator learns which text fragments could be candidate rationales and the encoder uses these candidates for prediction. Both the generator and the encoder are jointly trained during the optimization phase. The model explanation is provided by associating the model prediction with a set of critical input words.  

\cite{dong2017improving} focuses on providing intrinsic explanations for models on video captioning tasks. An interpretive loss function is defined to increase the visual fidelity of the learned features. This method is based on the nature of the used dataset, which contains rich human descriptions along with each video, and the rich text information can be leveraged to add constraint towards explainability. To produce an explanation, semantically meaningful concepts are first pre-extracted from human descriptions via Latent Dirichlet Allocation, which covers a variety of visual concepts such as objects, actions, relationships, etc. Based on the pre-extracted semantic topic, an interpretive loss is added to the original video captioning DNN model, for jointly training to generate video captions along with forcing the hidden neurons to be associated with semantic concepts.

From a practitioner's perspective, explanation association can be powerful because semantically meaningful concepts are directly interpretable by humans. Explanations based on these concepts can be represented in various forms, e.g., a relational graph or a heatmap, unlike text explanations which are limited to text only. Similar to text explanations, explanation association can also require specialized (labeled) datasets, which are difficult to obtain. Unlike text explanations, which mostly use an external component to generate the explanations, explanation associations often use internal model representations to associate high-level concepts. The practitioner may need to modify existing models to gain access to these internal representations, or in some cases, the internal representations might not be accessible to the practitioner. Furthermore, various methods in this category require a pipeline of several models where each model can become a bottleneck in the joint training process.
\newline

\noindent {\bf Model Prototype.}
This type of intrinsic method is specifically for classification tasks, and is derived from a classical form of case-based reasoning~\cite{kolodner1992introduction} called prototype classification~\cite{marchette2003classification,bien2011prototype,kim2014bayesian}. A prototype classifier generates classifications based on the similarity between the given input and each prototype observation in the dataset. In prototype classification applications, the word ``prototype'' is not limited to an observation in the dataset, but can be generalized to a combination of several observations or a latent representation learned in the feature space. To provide intrinsic explanations, the model architecture is designed to enable joint training the prototypes along with the original task. The model explainability is achieved by tracing the reasoning path for the given prediction back to each prototype learned by the model.

\cite{li2018deep} proposes an explainable prototype-based image classifier that can trace the model classification path to enable reasoning transparency. The model contains two major components; an autoencoder and a prototype classifier. The autoencoder, containing an encoder and a decoder, is to transform raw input into a latent feature space, and the latent feature is later used by the prototype classifier for classification. The prototype classifier, on the other hand, generates a classification via i) first calculating the distances in the latent space between a given input image and each prototype, ii) then passing through a fully-connected layer to compute the weighted sum of the distances, and iii) finally normalizing the weighted sums by the softmax layer to generate the classification result. Because the network learns \textit{prototypes} during the training phase, each prediction always has an explanation that is faithful to what the network actually computes. Each prototype can be visualized by the decoder, and the reasoning path of the prototype classifier can be partially traced given the fully-connected layer weights and the comparison between input and each visualized prototype, providing intrinsic model explanations.

\cite{chen2019looks} introduces an explainable DNN architecture called Prototypical Part Network (ProtoPNet) for image classification tasks. Similar to~\cite{li2018deep}, ProtoPNet also contains two components; a regular convolutional neural network and a prototype classifier.
The regular convolutional neural network projects the raw image into hidden feature space, where prototypes are learned. The prototype classifier is to generate model predictions based on the weighted sum of each similarity score between an image patch and a learned prototype.
Unlike~\cite{li2018deep} where learned prototypes are corresponding to the entire image, the prototypes in \cite{chen2019looks} are more fine-grained and are latent representations of parts/patches of the image. To provide a model explanation, the latent representation of each prototype is associated with an image patch in the training set, shedding light on the reasoning clue of ProtoPNet.

Compared to the other methods in this category, prototypes adopt a different approach towards creating interpretable model architecture. Other methods tend to use existing model architectures and make adaptations that either grant the practitioner access to either internal model components or add models to the existing model pipeline. In the case of prototypes, the practitioner often creates a novel architecture with traceable paths of reasoning. In some sense, model prototypes avoid end-to-end architectures where a single DNN learns the entire task. Instead, the DNN is constructed having explicitly interpretable components baked in as part of its design. 

\subsection{A Methods Lookup Table}
\label{section:explanation-methods-lookup-table}

Methods discussed in this field guide are categorized by distinct philosophies on eliciting and expressing an explanation from a DNN. 
This organization is ideal to understand the ``classes'' of methods that are being investigated in research and gradually implemented in practice. 
This does not, however, resolve an obvious question from a machine learning practitioner: 
\textit{What is the ``right'' type of explanatory method I should use when building a model to solve my specific kind of problem?}
It is difficult to match the methods with a particular situation because the type of explanation method suitable to that particular situation is often dependent on many variables including the type of DNN architecture, data, problem, and desired form of explanation. 

We propose Table~\ref{table:xai_lookup_table} and Table~\ref{table:intrinsic_lookup_table} as a starting point in answering this question. All of the papers in Figure~\ref{figure:methods} are organized in Table~\ref{table:xai_lookup_table} and Table~\ref{table:intrinsic_lookup_table}. Each table is titled with the main category of explanation method. Both tables are organized into five columns. The first column indicates the subcategory of the explanation method and the second column displays the reference to the explanation paper. The following three columns contain summarized information taken directly from the explanation paper. The third and fourth columns contain one or more icons representing the type of data used and the type of problem(s) presented in the paper respectively. The meaning of the icons can be found at the bottom of the table. The final column displays information about the specific DNN model on which the explanation method has been used in the paper.

The practitioner can make use of Table~\ref{table:xai_lookup_table} and Table~\ref{table:intrinsic_lookup_table} by considering what type of data, problem, and DNN architecture they are using in their specific situation. Then the practitioner can find an appropriate explanation method by matching the type of data, problem, and DNN architecture with the ones in the tables. For example, if a practitioner is using an image dataset to train a CNN on a classification problem, the practitioner can make use of all the explanation methods for which the \faImage icon and the \faCopyright[regular] icon and ``CNN'' are present in the respective rows. Note that in the ``DNN Type'' column we use ``no specific requirements'' to indicate that the DNN used in the respective paper does not need to meet any other specific requirements other than being a DNN. We use ``model agnostic'' to indicate that the type of model does not matter, i.e., the model does not have to be a DNN.

\begin{table}[p]
    \centering
    \scriptsize													
    \resizebox{\textwidth}{!}{													
    													
    \begin{tabular}{c|>{\tiny}p{3cm}|P{1.5cm}|P{2.1cm}|>{\tiny}p{3.8cm}}
    
    \multicolumn{5}{c}{\textbf{a) Visualization Methods}} \\													
    \specialrule{.2em}{.1em}{.1em}													
    & \multicolumn{1}{l|}{\textbf{Explanation Paper}} & \textbf{Data Type} & \textbf{Problem Type} & \multicolumn{1}{l}{\textbf{DNN Type}}\\													
    \specialrule{.1em}{.1em}{.1em}													
    													
    \parbox[t]{2mm}{\multirow{16}{*}{\rotatebox[origin=c]{90}{\textbf{Back-Propagation}}}}	
    
        &	\cellcolor{lightgray}	\cite{erhan2009visualizing}
        &	\cellcolor{lightgray}	\faImage	
        &	\cellcolor{lightgray}	\faCopyright[regular]	
        &	\cellcolor{lightgray}	classifier has to be differentiable	\\
        
    	&	\cite{zeiler2011adaptive}	
    	&	\faImage	
    	&	\faCopyright[regular]	
    	&	CNN with max-pooling + relu	\\
    	
    	&	\cellcolor{lightgray}	\cite{zeiler14}	
    	&	\cellcolor{lightgray}	\faImage	
    	&	\cellcolor{lightgray}	\faCopyright[regular]	
    	&	\cellcolor{lightgray}	CNN with max-pooling + relu	\\
    	
    	&	\cite{selvaraju2017grad}	
    	&	\faImage	
    	&	\faCopyright[regular]  \faHorse \faClosedCaptioning 
    	&	CNN	\\
    	
    	&	\cellcolor{lightgray}	\cite{zhou2016learning}
    	&	\cellcolor{lightgray}	\faImage \faNewspaper[regular]	
    	&	\cellcolor{lightgray}	\faCopyright[regular] \faRegistered[regular] \faHorse \faMapMarker*	
    	&	\cellcolor{lightgray}	CNN with global average pooling + softmax output\\
    	
    	&	\cite{bach15}	
    	&	\faImage \faNewspaper[regular]	
    	&	\faCopyright[regular]	
    	&	multilayer network	\\
    	
    	&	\cellcolor{lightgray}	\cite{lapuschkin2016analyzing}	
    	&	\cellcolor{lightgray}	\faImage	
    	&	\cellcolor{lightgray}	\faCopyright[regular]	
    	&	\cellcolor{lightgray}	CNN	\\
    	
    	&	\cite{arras2016explaining}	
    	&	\faBook	
    	&	\faCopyright[regular]	
    	&	CNN	\\
    	
    	&	\cellcolor{lightgray}	\cite{arras2017relevant}	
    	&	\cellcolor{lightgray}	\faBook	
    	&	\cellcolor{lightgray}	\faCopyright[regular]	
    	&	\cellcolor{lightgray}	CNN	\\
    	
    	&	\cite{ding2017visualizing}	
    	&	\faNewspaper[regular]	
    	&	\faArrowAltCircleRight	
    	&	attention-based encoder decoder	\\
    	
    	&	\cellcolor{lightgray}	\cite{montavon2017explaining}
    	&	\cellcolor{lightgray}	agnostic	
    	&	\cellcolor{lightgray}	\faCopyright[regular]	
    	&	\cellcolor{lightgray}	no specific requirements	\\
    	
    	&	\cite{shrikumar2017learning}	
    	&	\faImage \faDna	
    	&	\faCopyright[regular]	
    	&	CNN	\\
    	
    	&	\cellcolor{lightgray}   \cite{sundararajan2017axiomatic}	
    	&	\cellcolor{lightgray}	\faImage \faNewspaper[regular] \faShare*	
    	&	\cellcolor{lightgray}	\faCopyright[regular] \faArrowAltCircleRight	
    	&	\cellcolor{lightgray}	no specific requirements	\\
    	
    	&	\cite{sundararajan2016gradients}	
    	&	\faImage \faNewspaper[regular] \faShare*	
    	&	\faCopyright[regular]	
    	&	no specific requirements	\\
    \hline		
    
    \parbox[t]{2mm}{\multirow{6}{*}{\rotatebox[origin=c]{90}{\textbf{Perturbation}}}}	
    
        &	\cellcolor{lightgray}	\cite{zeiler14}	
        &	\cellcolor{lightgray}	\faImage	
        &	\cellcolor{lightgray}	\faCopyright[regular]	
        &	\cellcolor{lightgray}	CNN	\\
        
    	&	\cite{li2016understanding}	
    	&	\faBook	
    	&	\faCopyright[regular] \faTags	
    	&	no specific requirements	\\
    	
    	&	\cellcolor{lightgray}	\cite{fong2017interpretable}
    	&	\cellcolor{lightgray}	\faImage	
    	&	\cellcolor{lightgray}	\faCopyright[regular]	
    	&	\cellcolor{lightgray}	model agnostic	\\
    	
    	&	\cite{zintgraf2017visualizing}	
    	&	\faImage	
    	&	\faCopyright[regular]	
    	&	CNN	\\
    	
    	&	\cellcolor{lightgray}	\cite{robnik2008explaining}
    	&	\cellcolor{lightgray}	\faList	
    	&	\cellcolor{lightgray}	\faCopyright[regular]	
    	&	\cellcolor{lightgray}	models has to output probabilities	\\
    	
    	&	\cite{dabkowski2017real}	
    	&	\faImage	
    	&	\faCopyright[regular]	
    	&	classifier has to be differentiable	\\
    													
    \specialrule{.2em}{.1em}{.1em}		
    \multicolumn{5}{c}{}\\	
    \multicolumn{5}{c}{\textbf{b) Model Distillation}} \\
    \specialrule{.2em}{.1em}{.1em}													
    \parbox[t]{2mm}{\multirow{5}{*}{\rotatebox[origin=c]{90}{\textbf{Loc. Appr.}}}}	
    
        &	\cellcolor{lightgray}	\cite{ribeiro2016should}	
        &	\cellcolor{lightgray}	\faImage \faNewspaper[regular] \faList
        &	\cellcolor{lightgray}	\faCopyright[regular]	
        &	\cellcolor{lightgray}	model agnostic	\\
        
    	&	\cite{ribeiro2016nothing}	
    	&	\faImage \faNewspaper[regular] \faList
    	&	\faCopyright[regular]	
    	&	model agnostic	\\
    	
    	&	\cellcolor{lightgray}	\cite{ribeiro2018anchors}	
    	&	\cellcolor{lightgray}	\faImage \faNewspaper[regular] \faTable	
    	&	\cellcolor{lightgray}	\faCopyright[regular] \faAnchor \faEdit	
    	&	\cellcolor{lightgray}	model agnostic	\\
    	
    	&	\cite{elenberg2017streaming}	
    	&	\faImage	
    	&	\faCopyright[regular]	
    	&	model agnostic	\\
    	
    	&	\cellcolor{lightgray}	\cite{baehrens10}	
    	&	\cellcolor{lightgray}	\faImage \faBook \faList	
    	&	\cellcolor{lightgray}	\faCopyright[regular]	
    	&	\cellcolor{lightgray}	model agnostic	\\
    	
    	&	\cite{lundberg2017unified}	
    	&	\faImage \faNewspaper[regular] \faList
    	&	\faCopyright[regular]
    	&	model agnostic	\\
    	
    	&	\cellcolor{lightgray}	\cite{heskes2020causal}	
    	&	\cellcolor{lightgray}	\faImage \faNewspaper[regular] \faList
    	&	\cellcolor{lightgray}	\faCopyright[regular]
    	&	\cellcolor{lightgray}	model agnostic	\\
    \hline				
    
    \parbox[t]{2mm}{\multirow{8}{*}{\rotatebox[origin=c]{90}{\textbf{Model Translation}}}}	
        &	\cite{hou2020learning}	
        &	\faNewspaper[regular]	
        &	\faCopyright[regular]	
        &	RNN	\\
        
    	&	\cellcolor{lightgray}	\cite{murdoch2017automatic}
    	&	\cellcolor{lightgray}	\faNewspaper[regular]	
    	&	\cellcolor{lightgray}	\faCopyright[regular] \faCommentDots[regular]	&	\cellcolor{lightgray}	LSTM	\\
    	
    	&	\cite{harradon2018causal}	
    	&	\faImage	
    	&	\faCopyright[regular]	
    	&	CNN	\\
    	
    	&   \cellcolor{lightgray}	\cite{frosst2017distilling}
    	&	\cellcolor{lightgray}	\faImage	
    	&	\cellcolor{lightgray}   \faCopyright[regular]	
    	&   \cellcolor{lightgray}   CNN	\\
    	
    	&   \cite{zhang2019interpreting}	
    	&	\faImage	
    	&	\faCopyright[regular]	
    	&	CNN	\\
    	
    	&   \cellcolor{lightgray}   \cite{tan2018learning}
    	&	\cellcolor{lightgray}   \faTable	
    	&	\cellcolor{lightgray}   \faCopyright[regular] \faRegistered[regular]
    	&	\cellcolor{lightgray}   no specific requirements	\\
    	
    	&	\cite{zhang2017growing}	
    	&	\faImage	
    	&	\faCopyright[regular]	
    	&	CNN	\\
    	
    	&   \cellcolor{lightgray}   \cite{zhang2017interpreting}
    	&	\cellcolor{lightgray}   \faImage	
    	&	\cellcolor{lightgray}   \faCopyright[regular]	
    	&	\cellcolor{lightgray}   CNN, GANs	\\
    													
    \specialrule{.2em}{.1em}{.1em}									
    \multicolumn{5}{c}{}\\											
    
    \multicolumn{5}{c}{													
    \fbox{													
    \tiny													
    \begin{tabular}{cl|cl}													
    \multicolumn{2}{l|}{\textbf{Data Types}} & \multicolumn{2}{l}{\textbf{Problem Types}}\\													
    \hline													
    \Tstrut\Bstrut													
    \faImage & image & \faCopyright[regular] & classification\\													
    \faNewspaper[regular] & text & \faMapMarker* & localization\\													
    \faShare* & molecular graph & \faHorse  & visual question answering\\													
    \faDna & DNA sequence & \faRegistered[regular] & regression\\													
    \faBook & embedding & \faArrowAltCircleRight & language translation\\													
    \faList & categorical data & \faTags & sequence tagging\\													
    \faTable & tabular data & \faAnchor & structured prediction\\													
    & & \faEdit & text generation\\			
    & & \faCommentDots[regular] & question answering\\				
    & & \faClosedCaptioning & captioning													
    \end{tabular}													
    }													
    }													
    \end{tabular}	
    }													
    \caption{Lookup table for the (a) visualization and (b) model distillation methods.}				
    \label{table:xai_lookup_table}		
\end{table}																								
\begin{table}[p]													
    \centering													
    \scriptsize													
    \resizebox{\textwidth}{!}{													
    													
    \begin{tabular}{c|>{\tiny}p{2.9cm}|P{1.5cm}|P{2.1cm}|>{\tiny}p{3.2cm}}
    
    \multicolumn{5}{c}{\textbf{Intrinsic Methods}} \\													
    \specialrule{.2em}{.1em}{.1em}													
    & \multicolumn{1}{l|}{\textbf{Explanation Paper}} & \textbf{Data Type} & \textbf{Problem Type} & \multicolumn{1}{l}{\textbf{DNN Type}}\\													
    \specialrule{.1em}{.1em}{.1em}													
    													
    \parbox[t]{2mm}{\multirow{16}{*}{\rotatebox[origin=c]{90}{\textbf{Attention Mechanisms}}}}	
    
        &	\cellcolor{lightgray}	\cite{vaswani2017attention}
        &	\cellcolor{lightgray}	\faNewspaper[regular]	
        &	\cellcolor{lightgray}	\faArrowAltCircleRight	
        &	\cellcolor{lightgray}	transformer	\\
        
    	&	\cite{devlin2019bert}	
    	&	\faNewspaper[regular]	
    	&	\faFont	
    	&	transformer	\\
    	
    	&	\cellcolor{lightgray}	\cite{bahdanau2014neural}	
    	&	\cellcolor{lightgray}	\faNewspaper[regular]	
    	&	\cellcolor{lightgray}	\faArrowAltCircleRight	
    	&	\cellcolor{lightgray}	RNN encoder-decoder	\\
    	
    	&	\cite{luong2015effective}	
    	&	\faNewspaper[regular]	
    	&	\faArrowAltCircleRight	
    	&	stacking LSTM	\\
    	
    	&	\cellcolor{lightgray}	\cite{wang2016attention}	
    	&	\cellcolor{lightgray}	\faNewspaper[regular]	
    	&	\cellcolor{lightgray}	\faHeart[regular]	
    	&	\cellcolor{lightgray}	attention-based LSTM	\\
    	
    	&	\cite{letarte2018importance}	
    	&	\faNewspaper[regular]	
    	&	\faHeart[regular] \faCopyright[regular]	
    	&	self-attention network	\\
    	
    	&	\cellcolor{lightgray}	\cite{he2018effective}
    	&	\cellcolor{lightgray}	\faNewspaper[regular]	
    	&	\cellcolor{lightgray}	\faHeart[regular]	
    	&	\cellcolor{lightgray}	attention-based LSTM	\\
    	
    	&	\cite{teney2018tips}	
    	&	\faImage \faNewspaper[regular]	
    	&	\faHorse	
    	&	CNN + GRU combination	\\
    	
    	&	\cellcolor{lightgray}	\cite{mascharka2018transparency}	
    	&	\cellcolor{lightgray}	\faImage	
    	&	\cellcolor{lightgray}	\faHorse	
    	&	\cellcolor{lightgray}	various specialized modules	\\
    	
    	&	\cite{xie2019visual}	
    	&	\faImage	
    	&	\faCheckCircle	
    	&	various specialized modules	\\
    	
    	&	\cellcolor{lightgray}	\cite{park2016attentive}	
    	&	\cellcolor{lightgray}	\faImage \faNewspaper[regular]	
    	&	\cellcolor{lightgray}	\faHorse	
    	&	\cellcolor{lightgray}	various specialized modules	\\
    	
    	&	\cite{vinyals2015show}	
    	&	\faImage	
    	&	\faClosedCaptioning	
    	&	CNN + LSTM combination	\\
    	
    	&	\cellcolor{lightgray}	\cite{xu2015show}	
    	&	\cellcolor{lightgray}	\faImage	
    	&	\cellcolor{lightgray}	\faClosedCaptioning	
    	&	\cellcolor{lightgray}	CNN + RNN combination	\\
    	
    	&	\cite{antol2015vqa}
    	&	\faImage \faNewspaper[regular]	
    	&	\faHorse	
    	&	CNN + MLP, CNN + LSTM combinations	\\
    	
    	&	\cellcolor{lightgray}	\cite{goyal2017making}
    	&	\cellcolor{lightgray}	\faImage \faNewspaper[regular]	
    	&	\cellcolor{lightgray}	\faHorse	
    	&	\cellcolor{lightgray}	CNN + LSTM combination	\\
    	
    	&	\cite{anderson2018bottom}	
    	&	\faImage \faNewspaper[regular]	
    	&	\faHorse \faClosedCaptioning	
    	&	region proposal network + resnet combo, LSTM	\\
    	
    \hline	
    
    \parbox[t]{2mm}{\multirow{13}{*}{\rotatebox[origin=c]{90}{\textbf{Joint Training}}}}	
    
        &	\cellcolor{lightgray}	\cite{camburu2018snli}
        &	\cellcolor{lightgray}	\faNewspaper[regular]	
        &	\cellcolor{lightgray}	\faFont	
        &	\cellcolor{lightgray}	LSTM	\\
        
    	&	\cite{hind2019ted}
    	&	\faBook	
    	&	\faCopyright[regular]	
    	&	model agnostic	\\
    	
    	&	\cellcolor{lightgray}	\cite{hendricks2016generating}	
    	&	\cellcolor{lightgray}	\faImage	
    	&	\cellcolor{lightgray}	\faCopyright[regular]	
    	&	\cellcolor{lightgray}	CNN	\\
    	
    	&	\cite{zellers2019recognition}	
    	&	\faImage \faNewspaper[regular]	
    	&	\faLink	
    	&	recognition to cognition network	\\
    	
    	&	\cellcolor{lightgray}	\cite{liu2019towards}	
    	&	\cellcolor{lightgray}	\faNewspaper[regular]	
    	&	\cellcolor{lightgray}	\faCopyright[regular]	
    	&	\cellcolor{lightgray}	encoder-predictor	\\
    	
    	&	\cite{park2018multimodal}	
    	&	\faImage \faNewspaper[regular]	
    	&	\faCopyright[regular] \faHorse	
    	&	pointing and justification model	\\
    	
    	&	\cellcolor{lightgray}	\cite{kim2018textual}	
    	&	\cellcolor{lightgray}	\faImage	
    	&	\cellcolor{lightgray}	\faTools	
    	&	\cellcolor{lightgray}	CNN	\\
    	
    	&	\cite{lei2016rationalizing}	
    	&	\faNewspaper[regular]	
    	&	\faHeart[regular]	
    	&	encoder-generator	\\
    	
    	&	\cellcolor{lightgray}	\cite{melis2018towards}
    	&	\cellcolor{lightgray}	\faImage \faList \faTable	
    	&	\cellcolor{lightgray}	\faCopyright[regular]	
    	&	\cellcolor{lightgray}	self-explaining neural network	\\
    	
    	&	\cite{iyer2018transparency}	
    	&	\faImage	
    	&	\faRobot	
    	&	deep q-network	\\
    	
    	&	\cellcolor{lightgray}	\cite{dong2017improving}	
    	&	\cellcolor{lightgray}	\faVideo	
    	&	\cellcolor{lightgray}	\faClosedCaptioning	
    	&	\cellcolor{lightgray}	attentive encoder-decoder	\\
    	
    	&	\cite{li2018deep}	
    	&	\faImage	
    	&	\faCopyright[regular]	
    	&	autoencoder + prototype layer combination	\\
    	
    	&	\cellcolor{lightgray}	\cite{chen2019looks}	
    	&	\cellcolor{lightgray}	\faImage	
    	&	\cellcolor{lightgray}	\faCopyright[regular]	
    	&	\cellcolor{lightgray}	prototypical part network	\\
    													
    \specialrule{.2em}{.1em}{.1em}		
    \multicolumn{5}{c}{}\\	
    
    \multicolumn{5}{c}{													
    \fbox{													
    \tiny													
    \begin{tabular}{cl|cl}													
    \multicolumn{2}{l|}{\textbf{Data Types}} & \multicolumn{2}{l}{\textbf{Problem Types}}\\													
    \hline													
    \Tstrut\Bstrut													
    \faImage	&	image	&	\faCopyright[regular]	&	classification	\\						
    \faNewspaper[regular]	&	text	&	\faHorse	&	visual question answering	\\						
    \faBook	&	embedding	&	\faArrowAltCircleRight	&	language translation	\\						
    \faList	&	categorical data	&	\faClosedCaptioning	&	captioning	\\						
    \faTable	&	tabular data	&	\faHeart[regular]	&	sentiment analysis	\\						
    \faVideo	&	video	&	\faCheckCircle	&	visual  entailment	\\						
    	&		&	\faLink	&	visual commonsense reasoning	\\						
    	&		&	\faFont	&	language understanding	\\						
    	&		&	\faRobot	&	reinforcement learning	\\						
    	&		&	\faTools	&	control planning	\\						
    \end{tabular}													
    }								
    }													
    \end{tabular}													
    }													
    \caption{Lookup table for the intrinsic methods.}			
    \label{table:intrinsic_lookup_table}
\end{table}	

\section{Evaluating Explanations}
\label{section:evaluation}

There is a growing body of research on the objective comparison of explanation methods and their quality. It is important to be able to evaluate the factual quality of the generated explanation. Evidence suggests that when humans and AI collaborate, humans often make better decisions when the AI provides a correct explanation~\cite{ray2019can}. When the explanation is incorrect it can lead to bad outcomes~\cite{jacovi2020towards}. In addition, practitioners need to know whether they can trust the explanation that the methods return. It is well known that explanation methods are subject to misinterpretation, especially visual explanation  methods~\cite{kindermans2019reliability,nie2018theoretical,adebayo2018sanity,alvarez2018robustness}. 
There are concerns regarding the factual correctness of explanation methods such as deconvolution, guided backpropagation and LRP~\cite{kindermans2018learning}.
One main challenge for explanation evaluations is the lack of ground truth for most cases. In addition, the favorable evaluation metric may vary a lot according to the specific evaluation goal and oriented user groups. 

\subsection{What Makes a Good Explanation?}
The answer to this question depends on the user, the context of use, the type of model and data, and the desired explanation form. To address this need, literature has come up with various desiderata~\cite{ras2018explanation,robnik2018perturbation,carvalho2019machine,jacovi2020towards}. The traits \textit{fidelity}, \textit{consistency}, \textit{stability} and \textit{comprehensibility} are most commonly scrutinized and discussed. 

\subsection{Methods for Evaluating Explanation Methods and their Explanations}
Practically there are two main approaches for evaluating explanations. The first is to devise an objective metric or benchmark to evaluate the explanations without human intervention~\cite{samek2017evaluating,hooker2019benchmark,vu2019unified,adebayo2018sanity,alvarez2018robustness}. This approach has the benefit of being able to compare numerous explanation methods with each other. Using objective benchmarks we can investigate to what extent desiderata such as fidelity, consistency and stability are being satisfied. 
Given that visualization methods that produce heatmaps are a popular and intuitive type of explanation method, it has gained most of the attention in the subfield of evaluation explanation. Specifically, evaluating heatmaps generated for image classification networks is the focus of various evaluation work. There also exists a small body of work in evaluating explanations in NLP that we will discuss shortly~\cite{deyoung2019eraser}. 
The second approach is to let a human evaluate the explanations~\cite{prasad2020extent,hase2020evaluating,jesus2021can}. By using humans to evaluate the explanations, we can investigate to what degree the following desiderata are satisfied: clarity, parsimony, comprehensibility and importance.

\subsubsection{Evaluating Heatmaps}
Even though the following evaluations focus on heatmaps derived from image classifiers, they can also be applied to heatmaps of text in NLP explanations, e.g., LIME or SHAP.
\cite{samek2017evaluating} and variations~\cite{kindermans2018learning,petsiuk2018rise} introduce a perturbation-based method for evaluating the quality of heatmaps. Using the method, they compare the quality of heatmaps generated by sensitivity analysis~\cite{simonyan2013deep}, deconvolution~\cite{zeiler14} and LRP~\cite{bach15} by replacing the regions of the image that correspond to the location of the heatmap with randomly uniform data and checking how much the classification score changes. According to their metric, the more the classification score changes, the better the heatmap corresponds to class-discriminating features. Their results show that the heatmaps produced by LRP correspond better to the class features compared to heatmaps produced by sensitivity analysis and deconvolution. 

However, \cite{hooker2019benchmark} argues that the perturbation-based method violates the assumption that the training and evaluation data come from the same distribution. In response, \cite{hooker2019benchmark} proposes a benchmark for evaluating feature importance estimates in DNNs. Their benchmark is called ROAR: \textbf{R}em\textbf{O}ve \textbf{A}nd \textbf{R}etrain. The goal of ROAR is to determine whether the removal of important information caused classification degradation or whether the introduction of the so-called uninformative information caused the modified images to go out of distribution, thereby causing classification degradation. It replaces the fraction of pixels deemed important according to some heatmap with the channel mean, similar to the perturbation-based method~\cite{samek2017evaluating}. However, there is an important difference: to deal with the fallacy of introducing out of distribution images in the evaluation phase, ROAR applies similar modifications to all the images in both training and test set. The explanation method is applied to all the train and test set images to obtain a heatmap for each image. Then they remove the same percentage of the deemed important pixels from the image and replace it with the channel mean of that image. Finally, they train separate models on the modified data and evaluate the classification accuracy. If the accuracy of the re-trained models goes down, we can say with some certainty that the removed information in the modified image is indeed the cause of the classification degradation. Methods like \cite{kindermans2019reliability,kindermans2016investigating} investigate the reliability of heatmaps by modifying the input with information that does not change the classification result and checking how the heatmaps change as a result. They find that various visualization methods are vulnerable to input modification and return incorrect heatmaps as a result. The main conclusion is that many visualization methods are unreliable because they do not satisfy input invariance. 

In contrast to the previous methods, \cite{vu2019unified} suggests a metric to evaluate heatmaps based on perturbing regions that are not indicated as important. Their metric is called \textit{c}-Eval, where \textit{c} is a number that indicates how robust the classifier is to perturbations in regions deemed as not important by the explanation method. This method indirectly measures how accurate the heatmaps are: the larger \textit{c}, the more robust the classifier, the more accurate the explanation method is at identifying class-discriminating features. Using \textit{c}-Eval they compare various explanation methods and find that there is a significant difference in the quality of heatmaps produced by black-box methods (e.g., SHAP, LIME) compared to back-propagation based methods (e.g., LRP, DeepLIFT). 

In an alternative approach, \cite{adebayo2018sanity} proposes two sanity checks for evaluating the quality of heatmaps. The first is the \textit{model parameter randomization test}, and it compares heatmaps generated by a trained model with heatmaps generated by a randomly initialized model. If the outputs are similar, the explanation method is insensitive to model properties such as the weights. The second sanity check is the \textit{data randomization test}, and it compares heatmaps generated by a model trained on the original dataset with heatmaps generated by a model trained on a version of the dataset where all the labels have been randomly permuted. If the heatmaps are similar, it indicates that the explanation method is not dependent on the relationship between the data and the labels that exist in the original data. The distance between the heatmaps are measured using various similarity metrics. 

\subsubsection{Evaluating NLP Explanations}
The use of attention-based deep learning models for NLP have increased significantly~\cite{vaswani2017attention,brown2020language}. Attention has often been argued to be intrinsically interpretable, see Section~\ref{subsection:intrinsic_methods}. However, recent studies~\cite{jain2019attention,serrano2019attention,baan2019transformer} show that attention is not always interpretable and that attention does not always lead to insight into model prediction. \cite{jain2019attention} first raises the point that, while we make the assumption that attention is implicitly interpretable because directly it provides insight into which words are important, this assumption has never been formally evaluated. That is, the relationship between attention weights and model output is not clear. The experiments of their results show that correlation between feature importance measures, like heatmapping, and the learned attention weights is weak. The results suggest that the ability of attention modules to provide meaningful explanations into model prediction is questionable at best. \cite{serrano2019attention} manipulates the attention weights in trained models and analyzes the resulting difference in model predictions. Their findings are mixed: sometimes higher attention weights correlate to model predictions but not always. \cite{baan2019transformer} reveals that some attention heads tend to specialize towards interpretable parts of a document, but this ability does not generalize to all documents. Also the specializations are not consistent over differently initialized models.

To reconcile these conflicting views some studies~\cite{vashishth2019attention,wiegreffe2019attention} conduct experiments to find situations where attention can be used to gain insight into model prediction. \cite{vashishth2019attention} presents experiments over a range of NLP tasks justifying both observations. They identify conditions when the attention weights are interpretable and correlate with text heatmaps. Their results also reveal that attention weights are not interpretable when the input only has a single sequence by showing that in this situation the attention weights function as a gating unit. \cite{wiegreffe2019attention} provides a set of experiments that show that attention can be used to gain insight into model predictions. Their conclusion is that the results from \cite{jain2019attention} do not disprove that attention can serve as an explanation. 

\subsubsection{Using Humans to Evaluate Explanations}
Research in this explanation sub-field is still in its infancy and given that there is significant variation among people, contexts and their needs, the results of the papers in this section should not be taken as absolute. In contrast to the previous evaluation methods, this section addresses methods that are concerned with how interpretable explanations are to humans. This approach has the benefit of revealing to what extent the specific setting and explanation method is interpretable and useful to people who will use them. Research has revealed that there is still a big gap between the perceived and actual usefulness of explanations. 

Model interpretability can be understood as how easy it is for a human to predict the output of the model on new input based on past predictions. This concept is called \textit{simulatability}. It was found that LIME improves simulatability of models trained on tabular data~\cite{hase2020evaluating}, however, subjective ratings about the explanations did not predict how useful the explanations actually were. 

Another way to judge explanations using a human baseline is by investigating how much model explanations align with human explanations. In an investigation of model alignment of transformer models for Natural Language Inference (NLI) it was found that BERT-based transformer models score the highest on alignment~\cite{prasad2020extent}. It should also be noted that the number of parameters in the model lead to worse model alignment and that alignment was not predicted by accuracy on NLI tasks.

Sometimes explanations like LIME and SHAP can hurt user performance, albeit not by very much. In a recent study by \cite{jesus2021can} the accuracy and decision time was measured when participants need to make decisions. The accuracy of the participants was higher when only the basic data was given compared to when both the data and the explanation was given. However, the accuracy gap was not significant. The decision time significantly decreases when explanations are given. 

An evaluation of which factors in explanations make them human interpretable came to the conclusion that explanations might have more in common with design principles~\cite{lage2019evaluation}. One factor that drove the results was the complexity of the explanations. The paper further identifies that regularizers can be used to optimize for the interpretability of ML systems.

\section{Topics Associated with Explainability}
\label{section:topics_associated}

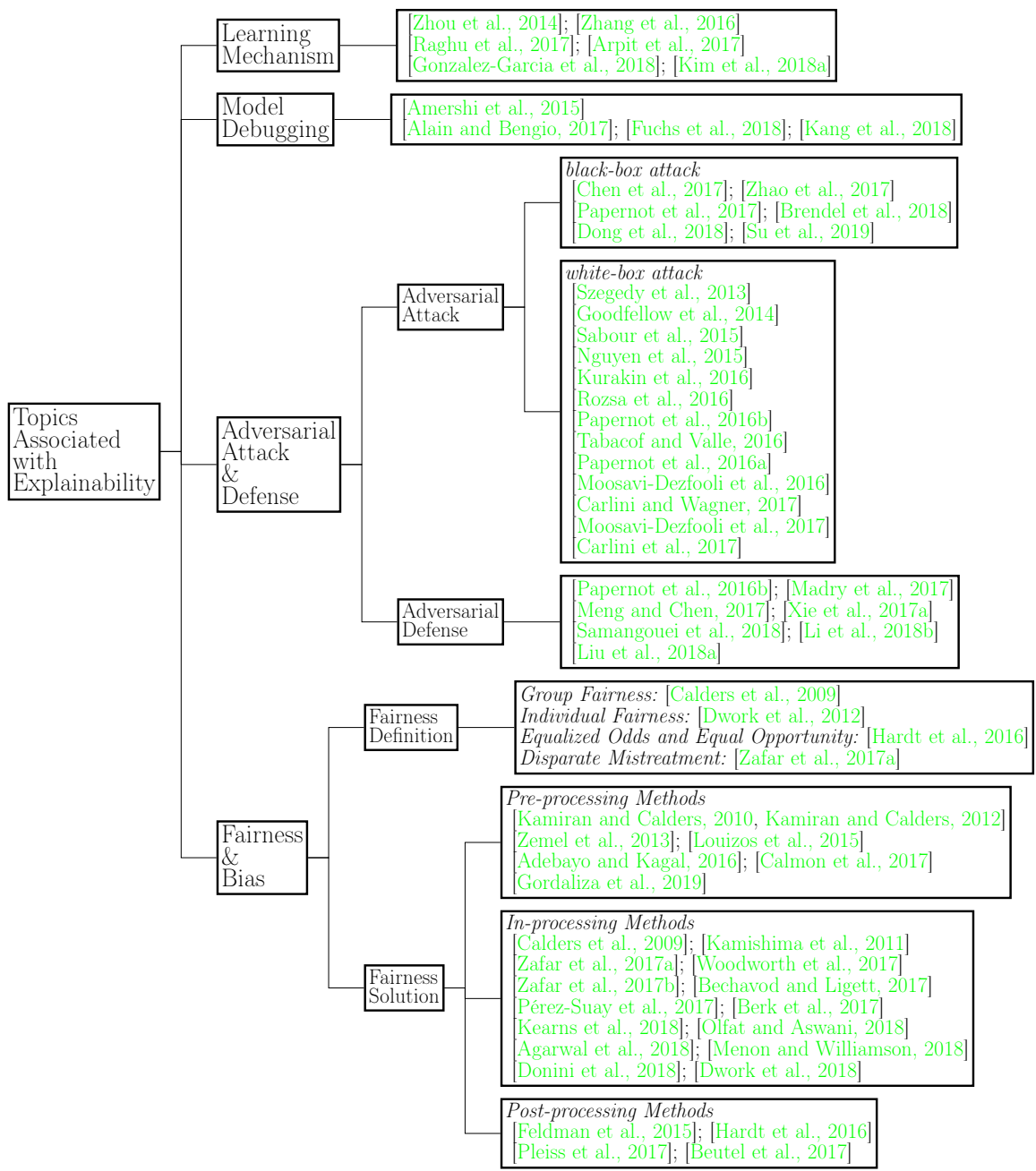
\begin{figure}[ht!]
    \begin{tabular}{c}
    \resizebox{\columnwidth}{!}{\begin{forest} 
    for tree={
    edge path={
          \noexpand\path[\forestoption{edge}](!u.parent anchor) -- +(20pt,0) |- (.child anchor)\forestoption{edge label};},
    l sep=5em, s sep=1em,
    child anchor=west,
    parent anchor=east,
    grow'=0,
    line width=0.75mm,
    anchor=west,
    draw,
    align=left,
    }
    [{{\Huge Topics}} \\ {{\Huge Associated}}\\
    {{\Huge with}} \\ {{\Huge Explainability}}\\
    [{{\Huge Learning}} \\ {{\Huge Mechanism}} 
    [             
    {{\huge \cite{zhou2014object};}} 
    {{\huge \cite{zhang2016understanding}}}\\
    {{\huge \cite{raghu2017svcca};}}
    {{\huge \cite{arpit2017closer}}}\\ 
    {{\huge \cite{gonzalez2018semantic};}}
    {{\huge \cite{kim2018interpretability}}}
    ]
    ]  
    [
    {{\Huge Model}} \\ {{\Huge Debugging}} \\
    [{{\huge \cite{amershi2015modeltracker}}}\\
    {{\huge \cite{alain16};}}
    {{\huge \cite{fuchs2018neural};}}
    {{\huge \cite{kang2018model}}}
    ]
    ]
    [{{\Huge Adversarial}}\\
    {{\Huge Attack}}\\
    {{\Huge \&}}\\
    {{\Huge Defense}}
    [{{\huge Adversarial}} \\ {{\huge Attack}} \\    
    [{{\huge \textit{black-box attack}}}\\
    {{\huge \cite{chen2017zoo};}}
    {{\huge \cite{zhao2017generating}}}\\
    {{\huge \cite{papernot2017practical};}}
    {{\huge \cite{brendel2017decision}}}\\
    {{\huge \cite{dong2018boosting};}}
    {{\huge \cite{su2019one}}}
    ]
    [
    {{\huge \textit{white-box attack}}}\\ 
    {{\huge \cite{szegedy2013intriguing}}} \\
    {{\huge \cite{goodfellow2014explaining}}} \\
    {{\huge \cite{sabour2015adversarial}}} \\
    {{\huge \cite{nguyen2015deep}}} \\
    {{\huge \cite{kurakin2016adversarial}}} \\
    {{\huge \cite{rozsa2016adversarial}}} \\
    {{\huge \cite{papernot2016distillation}}} \\
    {{\huge \cite{tabacof2016exploring}}} \\
    {{\huge \cite{papernot2016limitations}}}\\
    {{\huge \cite{moosavi2016deepfool}}}\\
    {{\huge \cite{carlini2017towards}}}\\
    {{\huge \cite{moosavi2017universal}}}\\
    {{\huge \cite{carlini2018ground}}}
    ]
    ]
    [{{\huge Adversarial}} \\ {{\huge Defense}} \\
    [{{\huge \cite{papernot2016distillation};}}
    {{\huge \cite{madry2017towards}}}\\
    {{\huge \cite{meng2017magnet};}}
    {{\huge \cite{xie2017mitigating}}}\\
    {{\huge \cite{samangouei2018defense};}}
    {{\huge \cite{li2018optimal}}}\\
    {{\huge \cite{liu2018adv}}}
    ]
    ]
    ]
    [{{\Huge Fairness}}\\
    {{\Huge \&}}\\
    {{\Huge Bias}}\\
    [{{\huge Fairness}} \\ {{\huge Definition}} \\
    [{{\huge \textit{Group Fairness:}~\cite{calders2009building}}}\\
    {{\huge \textit{Individual Fairness:}~\cite{dwork2012fairness}}}\\
    {{\huge \textit{Equalized Odds and Equal Opportunity:}~\cite{hardt2016equality}}}\\
    {{\huge \textit{Disparate Mistreatment:}~\cite{zafar2017fairness}}}
    ]
    ]
    [{{\huge Fairness}} \\ {{\huge Solution}} \\
    [{{\huge \textit{Pre-processing Methods}}}\\
    {{\huge \cite{kamiran2010classification,kamiran2012data}}}\\
    {{\huge \cite{zemel2013learning};}}
    {{\huge \cite{louizos2015variational}}}\\
    {{\huge \cite{adebayo2016iterative};}}
    {{\huge \cite{calmon2017optimized}}}\\
    {{\huge \cite{gordaliza2019obtaining}}}
    ]
    [{{\huge \textit{In-processing Methods}}}\\
    {{\huge \cite{calders2009building};}}
    {{\huge \cite{kamishima2011fairness}}}\\
    {{\huge \cite{zafar2017fairness};}}
    {{\huge \cite{woodworth2017learning}}}\\
    {{\huge \cite{zafar2017parity};}}
    {{\huge \cite{bechavod2017penalizing}}}\\
    {{\huge \cite{perez2017fair};}}
    {{\huge \cite{berk2017convex}}}\\
    {{\huge \cite{kearns2018preventing};}}
    {{\huge \cite{olfat2018spectral}}}\\
    {{\huge \cite{agarwal2018reductions};}}
    {{\huge \cite{menon2018cost}}}\\
    {{\huge \cite{donini2018empirical};}}
    {{\huge \cite{dwork2018decoupled}}}
    ]
    [{{\huge \textit{Post-processing Methods}}}\\
    {{\huge \cite{feldman2015certifying};}}
    {{\huge \cite{hardt2016equality}}}\\
    {{\huge \cite{pleiss2017fairness};}}
    {{\huge \cite{beutel2017data}}}
    ]
    ]
    ]
    ]
    \end{forest}} 
    \end{tabular}
    \caption{Topics associated with explainability.}
    \label{figure:topics_associated}
\end{figure}

We next review research topics closely aligned with explainable deep learning. 
A survey, visualized in Figure~\ref{figure:topics_associated}, identifies four broad related classes of research. Work on \textbf{learning mechanism (Section~\ref{subsection:learning_mechanism})} investigates the backpropagation process to establish a theory around weight training. 
These studies, in some respects, try to establish a theory to explain how and why DNNs converge to some decision-making process. Research on \textbf{model debugging (Section~\ref{subsection:model_debugging})} develops tools to recognize and understand the failure modes of a DNN. It emphasizes the discovery of problems that limit the training and inference process of a DNN (e.g., dead ReLUs, mode collapse, etc.). Techniques for \textbf{adversarial attack and defense (Section~\ref{subsection:adversarial_examples})} search for differences between regular and unexpected activation patterns. This line of work promotes deep learning systems that are robust and trustworthy; traits that also apply to explainability. Research on \textbf{fairness and bias in DNNs (Section~\ref{subsection:fairness_and_bias})} is related to the ethics trait discussed above, but more narrowly concentrates on ensuring DNN decisions do not over-emphasize undesirable input data features. We elaborate on the connection between these research areas and explainable DNNs next.

\subsection{Learning Mechanism}
\label{subsection:learning_mechanism}

The investigation of the learning mechanism tries to derive principles explaining the evolution of a model's parameters during training. Many existing approaches can be categorized as being semantics-related, in that the analysis tries to associate a model's learning process with concepts that have a concrete semantic meaning. They generally assign semantic concepts to a DNNs' internal filters (weights) or representations (activations), in order to uncover a human-interpretable explanation of the learning mechanism. Semantically interpretable descriptions are rooted in the field of neuro-symbolic computing~\cite{garcez2012neural}. An early work is~\cite{zhou2014object} which assigns semantic concepts, such as objects, object parts, etc, to the internal filters of a convolutional neural network (CNN) image scene classifier. Those semantic concepts are generated based on the visualization of receptive fields of each internal unit in the given layers. The authors also discovered that object detectors are embedded in a scene classifier without explicit object-level supervision for model training. \cite{gonzalez2018semantic} further explores this problem in a quantitative fashion. Two quantitative evaluations are conducted to study whether the internal representations of CNNs really capture semantic concepts. Interestingly, the authors' experimental results show that the association between internal filters and semantic concepts is modest and weak. But this association improves for deeper layers of the network, matching the conclusion of~\cite{zhou2014object}. \cite{kim2018interpretability} quantifies the importance of a given semantic concept with respect to a classification result via Testing with Concept Activation Vector (TCAV), which is based on multiple linear classifiers built with internal activations on prepared examples. The prepared examples contain both positive examples representing a semantic concept and randomly sampled negative examples that do not represent the concept. Directional derivatives are used to calculate TCAV, which measures the proportion of examples that belong to a given class that are positively influenced by a given concept. 

Other methods to interpret the learning process of a DNN searches for statistical patterns indicative of convergence to a learned state. Those learning patterns include but are not limited to:
i) how layers evolve along with the training process~\cite{raghu2017svcca};
ii) the convergence of different layers~\cite{raghu2017svcca}; and
iii) the generalization and memorization properties of DNNs~\cite{zhang2016understanding,arpit2017closer}. 
In studying the learning dynamics during training, \cite{raghu2017svcca} makes a comparison between two different layers or networks via Singular Vector Canonical Correlation Analysis (SVCCA). For a neuron in a selected layer of a DNN, the neuron's vector representation is generated in a ``global fashion'', i.e. all examples from a given finite dataset are used, and each element in the neuron's vector representation is an activation for an example. The vector representations for all neurons in a selected layer form a vector set, representing this layer. To compare two layers, SVCCA takes the vector set of each layer as input and calculates a canonical correlation similarity to make the alignment. The nature of SVCCA makes it a useful tool to monitor how layer activations evolve along with the training process. The authors further discover that earlier layers converge faster than later layers. Thus, the weights for earlier layers can be frozen earlier to reduce computational cost during training. Layer-wise convergence is also studied in work such as \cite{zhang2016understanding} using systematic experimentation. Keeping the model structure and hyper-parameters fixed, the authors' experiments are conducted only with different input modification settings, either on input labels or image pixels. The experimental results indicate that DNNs can perfectly fit training data with both random feature values and labels, while the degree of generalization on testing data reduces as randomness increases. The authors also hypothesize that explicit regularization (such as dropout, weight decay, data augmentation, etc.) \textit{may} improve generalization and stochastic gradient descent could act as an implicit regularizer for linear models. In a similar study, \cite{arpit2017closer} examines memorization by DNNs via quantitative experiments with real and random data. The study finds that DNNs do not simply memorize all real data; instead, patterns that are commonly shared among the data are leveraged for memorization. Interestingly, the authors claim that explicit regularization does make a difference in the speed of memorization for random data, which is different from the conclusions in~\cite{zhang2016understanding}. Besides the aforementioned research work, we would like to refer readers to a recent review paper~\cite{bahri2020statistical}, which covers the intersection between statistical mechanics and deep learning, and derives the success of deep learning from a theoretical perspective.

\subsection{Model Debugging} 
\label{subsection:model_debugging}

Similar to the concept of software debugging, the concept of model debugging applies techniques from traditional programming to find out when and where model-architecture, data processing, and training related errors occur. A ``probe'' is leveraged to analyze the internal pattern of a DNN, to provide further hints towards performance improvement. A probe is usually an auxiliary model or a structure such as a linear classifier, a parallel branch of the model pipeline, etc. The training process of the probe is usually independent of the training process of the master model (a DNN) that the probe serves for. Regardless of the form of the probe, the ultimate goal is model improvement. 

\cite{kang2018model} uses \textit{model assertions}, or Boolean functions, to verify the state of the model during training and run time. The assertions can be used to ensure the model output is consistent with meta observations about the input. For example, if a model is detecting cars in a video, the cars should not disappear and reappear in successive frames of the video. Model debugging is thus implemented as a verification system surrounding the model and is implicitly model-agnostic. The model assertions are implemented as user-defined functions that operate on a recent history of the model input and output. The authors explore several ways that model assertions can be used during both run-time and training time, in correcting wrong outputs and in collecting more samples to perform active learning. \cite{amershi2015modeltracker} proposes \textit{ModelTracker}, a debugging framework revolving around an interactive visual interface. This visual interface summarizes traditional summary statistics, such as AUC and confusion matrices, and presents this summary to the user together with a visualization of how close data samples are to each other in the feature space. The interface also has an option to directly inspect prediction outliers in the form of the raw data with its respective label, giving users the ability to directly correct mislabeled samples. The goal of this framework is to provide a unified, model-agnostic, inspection tool that supports debugging of three specific types of errors: mislabeled data, inadequate features to distinguish between concepts and insufficient data for generalizing from existing examples. \cite{alain16} uses linear classifiers to understand the predictive power of representations learned by intermediate layers of a DNN. The features extracted by an intermediate layer of a deep classifier are fed as input to the linear classifier. The linear classifier has to predict which class the given input belongs to. The experimental results show that the performance of the linear classifier improves when making predictions using features from deeper layers, i.e., layers close to the final layer. This suggests that task-specific representations are encoded in the deeper layers. \cite{fuchs2018neural} proposes the idea of \textit{neural stethoscopes}, which is a general-purpose framework used to analyze the DNN learning process by quantifying the importance of specific influential factors in the DNN and influence the DNN learning process by actively promoting and suppressing information. Neural stethoscopes extend a DNN's architecture with a parallel branch containing a two-layer perceptron. It is important to note that the main network branch does not need to be changed to be able to use the neural stethoscope. This parallel branch takes the feature representation from an arbitrary layer from the main network as input and is trained on a supplemental task given known complementary information about the dataset. Specifically, in this study the experiments are conducted on the ShapeStacks dataset~\cite{groth2018shapestacks}, which introduces a vision-based stability prediction task for block towers. The dataset provides information on both the local and global stability of a stack of blocks. In this specific study the stethoscope was used to investigate the internal representations contained in the network layers that lead to the prediction of the global stability of a stack of blocks, with local stability as complementary information. The stethoscope can be tuned to three different modes of operation: analytic, auxiliary, and adversarial. Each mode determines how the stethoscope loss $L_S$ is propagated, e.g., in the analytical mode, $L_S$ is not propagated through the main network. The auxiliary and adversarial modes are used to promote and suppress information respectively. The paper shows that the method was successful in improving network performance and mitigating biases that are present in the dataset. 

\subsection{Adversarial Attack and Defense}
\label{subsection:adversarial_examples}

An adversarial example is an artificial input engineered to intentionally disturb the judgment of a DNN~\cite{goodfellow2014explaining}. Developing defenses to adversarial examples requires a basic understanding of the space that inputs are taken from and the shape and form of boundaries between classes. Interpretations of this space inform the construction of defenses to better discriminate between classes and forms the basis of explaining input/output behavior. Moreover, an ``explanation'' from a model that is not reasonable given its input and output may be indicative of an adversarial example. 

The study of adversarial examples~\cite{yuan2019adversarial,zhang2019generating} are from the perspective of attack and defense. {\em Adversarial attack} methods are about generating adversarial examples that can fool a DNN. From the model access perspective, there are two main types of adversarial attack: \textit{black-box }~\cite{chen2017zoo,zhao2017generating,papernot2017practical,brendel2017decision,dong2018boosting,su2019one} and \textit{white-box}~\cite{szegedy2013intriguing,goodfellow2014explaining,sabour2015adversarial,nguyen2015deep,kurakin2016adversarial,rozsa2016adversarial,papernot2016limitations,moosavi2016deepfool,tabacof2016exploring,kurakin2016adversarial,carlini2017towards,moosavi2017universal,carlini2018ground,eykholt2018robust} attacks. In the black-box setting the attacker has no access to the model parameters or intermediate gradients whereas these are available for the white-box settings. {\em Adversarial defense}~\cite{madry2017towards,papernot2016distillation,meng2017magnet,xie2017mitigating,samangouei2018defense,li2018optimal,liu2018adv}, on the other hand, seeks solutions to make a DNN robust against generated adversarial examples. 

Recent work on adversarial attack reveals vulnerabilities by perturbing input data with imperceptible noise~\cite{goodfellow2014explaining,carlini2017towards,madry2017towards} or by adding ``physical perturbations'' to objects under analysis (i.e. black and white stickers on objects captured by computer vision systems)~\cite{eykholt2018robust}. Among numerous adversarial attack methods, the C\&W attack~\cite{carlini2017towards} and PGD attack~\cite{madry2017towards} are frequently used to evaluate the robustness of DNNs. 

C\&W attack~\cite{carlini2017towards} casts the adversarial attack task as an optimization problem and is originally proposed to challenge an adversarial defense method called defensive distillation~\cite{papernot2016distillation}. Variants of C\&W attacks are based on the distance metrics ($\ell_0$, $\ell_2$, or $\ell_\infty$). \cite{carlini2017towards}, for example, can successfully defeat defensive distillation with high-confidence adversarial examples generated via C\&W attack. Projected Gradient Descent (PGD) attack~\cite{madry2017towards} in brief is an iterative version of an early stage adversarial attack called Fast Gradient Sign Method (FGSM)~\cite{goodfellow2014explaining}. As indicated in its name, PGD attack generates adversarial examples based on the gradients of the loss with respect to the input. PGD attack is more favorable than C\&W attack when direct control of input distortion is needed~\cite{liu2018adv}.

Adversarial defense is challenging due to the diversity of the adversarial example crafting processes and a DNN's high-dimensional feature space. There exist two typical groups of adversarial defense methods, 
i) adversarial training~\cite{madry2017towards,goodfellow2014explaining,szegedy2013intriguing}, which is to augment the training dataset with generated adversarial examples such that the trained model is more robust against adversarial attack, and
ii) removal perturbations~\cite{samangouei2018defense,meng2017magnet}, which dismisses adversarial perturbations from input data.
\cite{madry2017towards} integrates the PGD attack into the model training process, such that the model is optimized on both benign examples and challenging adversarial examples. The optimization is conducted in a min-max fashion, where the loss for adversarial attack process is maximized in order to generate strong adversarial examples, while the loss for the classification process is minimized, in order to get a robust and well-performed model. \cite{samangouei2018defense}, on the other hand, tackles the adversarial defense problem by filtering out adversarial perturbations. Generative Adversarial Networks (GANs) are leveraged to project a given input image, potentially polluted by adversarial perturbations, into a pseudo original image, where adversarial artifacts are diminished. Model decisions are made from the GAN generated ``original’’ image. Experiments indicate this defense technique is effective against both black-box and white-box attacks.

\subsection{Fairness and Bias} 
\label{subsection:fairness_and_bias}

Model fairness aims to build DNN models that objectively consider each input feature and is not unduly biased against a particular subset of the input data. Although a firm definition of what it means for a DNN to be ``fair'' is evolving, common themes are emerging in the literature~\cite{heidari2018fairness}. \textit{Group fairness}~\cite{calders2009building}, also called \textit{demographic parity} or \textit{statistical parity}, focuses on fairness with respect to a group (based on race, gender, etc.). The goal of group fairness is to ensure each group receives equalized percentage of benefit. Consider a loan application as an example. Suppose we are monitoring the loan approval situation of two cities, city A and city B. The population of city A is twice as much as that of city B. Based on the definition of group fairness, twice as many loan applications should be approved in A compared to city B. \textit{Individual fairness}~\cite{dwork2012fairness} aims to treat similar inputs similarly based on a metric to measure the closeness of their features. To compare group fairness and individual fairness, let's return to the loan request example. Under the restriction of group fairness, an individual from city A may not be approved for a loan request just because of the group percentage limitation, even though this individual is more qualified based on economic metrics than other approved ones from city B. However, individual fairness requires that individuals with similar characteristics should have the same chance to be approved for a loan request, regardless of which city individuals come from. This is in antithesis with group fairness. Further notions of fairness, such as \textit{equalized odds} and \textit{equal opportunity}~\cite{hardt2016equality}, \textit{disparate mistreatment}~\cite{zafar2017fairness}, and others~\cite{heidari2018fairness,woodworth2017learning}
are also studied in the literature. 

The fairness problem is currently addressed by three types of methods~\cite{calmon2017optimized}:
(i) {\em pre-processing} methods revise input data to remove information correlated to sensitive attributes;
(ii) {\em in-process} methods add fairness constraints into the model learning process;
and~(iii) {\em post-process} methods adjust model predictions after the model is trained. \textit{Pre-processing} methods~\cite{kamiran2010classification,kamiran2012data,zemel2013learning,louizos2015variational,adebayo2016iterative,calmon2017optimized,gordaliza2019obtaining}  learn an alternative representation of the input data that removes information correlated to the sensitive attributes (such as race or gender) while maintaining the model performance as much as possible. For example, \cite{calmon2017optimized} proposes a probabilistic framework to transform input data to prevent unfairness in the scope of supervised learning. The input transformation is conducted as an optimization problem, aiming to balance discrimination control (group fairness), individual distortion (individual fairness), and data utility. \textit{In-process} methods~\cite{calders2009building,kamishima2011fairness,zafar2017fairness,woodworth2017learning,zafar2017parity,bechavod2017penalizing,kearns2018preventing,perez2017fair,berk2017convex,olfat2018spectral,agarwal2018reductions,menon2018cost,donini2018empirical,dwork2018decoupled} directly introduce fairness learning constraints to the model in order to punish unfair decisions during training. \cite{kamishima2011fairness} achieves the fairness goal by adding a fairness regularizer, for example, such that the influence of sensitive information on model decisions is reduced. \textit{Post-process} methods~\cite{feldman2015certifying,hardt2016equality,pleiss2017fairness,beutel2017data} are characterized by adding ad-hoc fairness procedures to a trained model. One example is \cite{hardt2016equality} which constructs non-discriminating predictors as a post-processing step to achieve equalized odds and equal opportunity (two fairness notions proposed in their study). They introduce the procedure to construct non-discriminating predictors for two scenarios of the original model, binary predictor and score function, where in the latter scenario the original model generates real score values in range $[0, 1]$. A non-discriminating predictor is constructed for each protected group, with a defined threshold to achieve a fairness goal.

\section{Designing Explanations for Users}
\label{section:designing}

The foundations of explaining DNNs discussed in this survey are seldom enough to achieve explanations useful to users in practice. ML engineers designing explainable DNNs in practice will thus often integrate an explanatory method into their DNN and then refine the presentation of the explanation to a form useful for the end-user. A {\em useful} explanation must conform to some definition of what constitutes a satisfactory explanation of the network's inner workings depending on the user, the conditions of use, and the task at hand. These definitions are often qualitative (e.g., one user is better swayed by visual over textual explanations for a task). User requirements for an explanation may further vary by preferences between explanations that are of high fidelity versus those that are parsimonious. The quality of an explanation depends on user- and context-specific utility and makes the evaluation of explanations a difficult problem.

This suggests that explanations, grounded in the methods discussed in this field guide, need to be designed by engineers on a case-by-case basis for the user and task at hand. This section describes the following important design questions when engineers apply the methods in this field guide in practice: 

\begin{enumerate}
    \item {\bf Who is the end user?} 
    The kind of end-user, and in particular their expertise in deep learning and their domain-specific requirements, define the appropriate trade-off between fidelity and parsimony in an explanation's presentation.
    \item {\bf How practically impactful are the decisions of the DNN?} 
    Here impact corresponds to the consequence of right and wrong decisions on people and society. 
    Time-critical scenarios require explanations that can be rapidly generated and processed by a user should there be a need to intervene (e.g., in self-driving cars). 
    Decision-critical scenarios require explanations that are {\em trustworthy}, that is, an explanation that a user trusts to be faithful to the actual decision-making process of the DNN. 
    \item {\bf How extendable is an explanation?} 
    It is expensive to design a form of explanation for only a single type of user who faces a single type of problem. 
    A good design should thus be grounded on a single user's preferences that can be applied to multiple types of problems, or be flexible enough to appeal to multiple user types examining the same problem type. 
    It may not be feasible to devise the presentation of an explanation that appeals to a broad set of users tailed to a diverse set of problems.
\end{enumerate}

\subsection{Understanding the End User}
\label{subsection:end_user}

One of the primary tasks to design an explanation is to determine the type of end-user using the system. The literature has documented cases of designs that provide both low-level technical specific explanations targeting on deep learning experts~\cite{zeiler14,sundararajan2017axiomatic,anderson2018bottom,li2016understanding,fong2017interpretable,zintgraf2017visualizing}, and high-level reasoning extracted explanations catering normal users~\cite{harradon2018causal,zhang2019interpreting,zhang2017growing,zhang2017interpreting}. DNN experts care mostly about technical details and potential hints for model revising and performance improvement. Ideal explanations for them could be in form of input feature influence analytics~\cite{adler2018auditing,koh2017understanding,li2016understanding,fong2017interpretable}, hidden states interaction and visualizations~\cite{anderson2018bottom,vaswani2017attention,vinyals2015show,zeiler14,selvaraju2017grad,bach15,sundararajan2017axiomatic}, etc. DNN experts, for instance, could check if the model is emphasizing reasonable image areas~\cite{zeiler14} or text elements/words~\cite{vaswani2017attention,sundararajan2017axiomatic} towards generating corresponding model decisions, and propose model revision strategies accordingly. Normal users, on the other hand, mainly focus on the high-level functionality of the model, instead of technical details. Their main concern is if the model is working reasonably and not violating human logic. The explanation can be represented in the form of extracted reasoning logic~\cite{harradon2018causal,zhang2019interpreting,zhang2017growing,zhang2017interpreting} or some easy to understandable input clues with respect to given prediction~\cite{ribeiro2016should}. If the decision is generated on unexpected input elements or not following a logical reasoning process, a doubt could be raised by the user to deny such a model decision. Considering the different user expertise level on DNN knowledge, the designing of a general model explanation system, which satisfies both DNN experts and normal users, is challenging and remains to be explored.

The domain a user operates in is another important consideration. For example, the explanation needs of a medical doctor require that the explanation representation be detailed enough such that the doctor can understand the reasoning process behind the specific diagnosis and be confident about said diagnosis~\cite{lipton2017doctor}, e.g., the patient needs this specific treatment because it identifies features of cancer at a particular stage. But no explainable method is able to automatically tailor its explanations to end-users for a specific domain. One possible way to obtain such explanations using the present art is if the features of the input data that are expressed by an explanation method have an intuitive domain-specific interpretation, which is built upon a systematic knowledge base constructed by domain experts.

\subsection{The Impact of DNN Decisions}
\label{subsection:critical Scenarios}

The need for and characteristics of an explanation depends on the impact of a DNN's operation on human life and society. This impact can be realized based on the impact and speed of a decision. In \textit{time-critical} scenarios~\cite{grigorescu2019survey} where users must process and react to DNN decisions in limited time, explanations must be produced that are simple to interpret and understand and are not computationally intense to perform. This is a particularly important aspect of explanations that is seldom investigated in the literature. For example, a DNN providing recommendations during a military operation, or sensing upcoming hazards on a vehicle, needs to support their output with explanations while giving the user enough time to process and react accordingly. In a \textit{decision-critical} scenario~\cite{grigorescu2019survey,nemati2018interpretable,ahmad2018interpretable}, the ability to not only interpret but deeply inspect a decision grows in importance. Any user decision based on a DNN's recommendation should be supported with evidence and explanations others can understand. At the same time, should a DNN's recommendation turn out to be incorrect or lead to an undesirable outcome for the user, the model should be inspected post-hoc to hypothesize root causes and identify ``bugs'' in the DNN's actions. Deep, technical inspections of the neural network guided by comprehensive interpretations of its inference and training actions are necessary for such post-hoc analysis. 

Few current model explanations are designed with time- and decision-critical scenarios in mind. The computational cost of many model explanations tends to be high and may require extra human labor, which is undesirable if an automatic instant explanation is needed. For instance, for explanations presented in form of model visualization~\cite{zeiler14,selvaraju2017grad,shrikumar2017learning,sundararajan2017axiomatic,montavon2017explaining,zhou2016learning}, extra human effort is needed for verification, which is potentially costly. Besides, some explanation methods with post-hoc training involved ~\cite{ribeiro2016should,frosst2017distilling,krakovna16,hou2020learning} may be limited in its utility on providing explanations for real-time input. The study for decision-critical scenarios is still under development. In order to increase the fidelity and reliability of model decisions, a variety of topics are explored besides model explanations, including model robustness~\cite{papernot2016distillation,meng2017magnet,xie2017mitigating,samangouei2018defense,li2018optimal}, fairness and bias~\cite{heidari2018fairness,calders2009building,hardt2016equality,zafar2017fairness,calmon2017optimized,gordaliza2019obtaining,agarwal2018reductions,menon2018cost,donini2018empirical,dwork2018decoupled,pleiss2017fairness,beutel2017data}, model trustworthiness~\cite{jiang2018trust,heo2018uncertainty}, etc. The study of the aforementioned topics, together with model explanations, may jointly shed light on potential new solutions for applications on decision-critical scenarios.

\subsection{Design Extendability}
\label{subsection:modularity_and_reusability}

Modularity and reusability are important extendability traits in the architecture of large-scale software systems: modularity promotes the ability of an engineer to replace and alter system components as necessary, while reusability promotes the use of already proven software modules. In a similar vein, highly reliable and performant DNN systems should also be constructed with reusable and highly modular components. Modularity is a trait of the functional units of a DNN that may be adaptable for multiple architectures. For example, the form of an attention mechanism suitable for any kind of sequential data processing~\cite{vaswani2017attention,devlin2019bert}. A highly modular DNN architecture may be one that contains many ``plug and play'' components in each layer so that its complete design can be seen as a composition of interconnected functional units. Reusability speaks to complete DNN systems, perhaps already trained, that can be reused in multiple problem domains. One example of reusability is the common application of a pre-trained YOLO~\cite{redmon2016you} model for object localization in frames in a deep learning video processing pipeline. 

DNN explanation methods that also exhibit these extendability traits are more likely to be broadly useful over a variety of DNN models and application domains. Modularized explainable models will crucially reduce the overhead in implementing and deploying explainability in new domains and may lead to explanatory forms a user is familiar with over multiple types of models. Reusability plays a role in risk control, such that the fidelity of an explanation remains consistent however the explanatory method is applied. 

Neither modularity nor reusability is the focus of explainable methods in the literature. However, existing methods could be divided by how modular they potentially are. Model-agnostic methods~\cite{ribeiro2016should,ribeiro2016nothing,ribeiro2016model,fong2017interpretable,jha2017learning}, which do not take the type of model into account, are modular by definition in the sense that the explanatory module is independent of the model it is producing explanations for. On the other hand, The second category contains explanation methods that are very specific to the model~\cite{shrikumar2017learning,zeiler14,bach15,montavon2017explaining,zhou2016learning,murdoch2018beyond,xie2017relating,park2018multimodal,hendricks2016generating}. This aspect is important for expert users that are developing deep learning models and need to understand specifically which aspect of the deep learning model is influencing the predictions, e.g. in model debugging. However, these methods by their very nature lack modularity. 

\section{Future Directions}
\label{section:future}

The field guide concludes by introducing research directions whose developments can contribute to improving explainable deep learning.
\newline

\textbf{A Unifying Approach to Explainability}. There have been several efforts to come up with a framework for explainable artificial intelligence (XAI) or interpretable machine learning~\cite{gilpin2018explaining,doshi2017towards,DoshiVelez2018ConsiderationsFE}. Each one of these papers considers the literature from a different perspective, with little consideration to unify various methods. Currently, there is still a lack of systematic general theory in the realm of DNN explanation~\cite{arrieta2019explainable,diez2013general}. A systematic theory should be able to benefit the overall model explanation studies, and once formed, some current challenging explainable problems may be able to be handled properly by nature, and some novel directions may be proposed based on the systematic theories. However, one of the main reasons why it is so difficult to establish a formal theory of explanation is that the basic concepts of explanations in AI are difficult or impossible to formalize~\cite{wolf2019formal}.
\newline

\noindent \textbf{User-friendly Explanations.}
User-friendly explanations are needed to minimize the technical understanding of a user to correctly interpret explanations. As the concern of the opaque nature of DNNs is raising increasing attention in the society and even required by law, model explanations would inevitably be mandatory in a wide range of real-life applications~\cite{goodman2017european}. Given the varied backgrounds of model users, the friendliness would be a future trend towards constructing explanations of high qualities. Most explainable methods are still catering towards expert users instead of laymen~\cite{ras2018explanation}, in the sense that knowledge about the method, for instance the DNN process, is needed to understand the explanation. The requirement on model knowledge limits the wide usage of such explanation models, since in real scenarios the chance of the end-users being machine learning experts is very low. Assuming the end-user has been correctly determined, the next step is to determine \textit{what} needs to be explained, i.e., which step or decision does the system need to explain?
\newline

\noindent \textbf{Producing Explanations Efficiently.}
Time and decision-critical explanations~\cite{grigorescu2019survey,ahmad2018interpretable,nemati2018interpretable}, as discussed in Section~\ref{subsection:critical Scenarios}, must be produced with enough time for a user to react to a DNN's decision. An efficient manner to produce explanations further saves computational power, which is favorable in industrial applications or when explanations are required in environments with low computing resources.
\newline

\noindent \textbf{Developing Methods for Trustworthiness.}
The vulnerability of a DNN to adversarial examples~\cite{yuan2019adversarial,goodfellow2014explaining,carlini2017towards,madry2017towards} and poisoned training sets~\cite{saha2019hidden} raises much concern on trustworthiness. As more and more DNNs are leveraged in real-life applications, the demand for model trustworthiness would undoubtedly increase, especially for decision-critical scenarios where undesired decisions may cost severe consequences. This thread of research is only beginning to be developed~\cite{jiang2018trust,heo2018uncertainty}. 

\section{Conclusions} 
\label{section:conclusions}

The rapid advancements in deep neural networks have stimulated innovations in a wide range of applications such as facial recognition~\cite{masi2018deep}, voice assistance~\cite{tulshan2018survey}, driving system~\cite{jain2015car}, etc. The field of deep learning explainability is motivated by the opaque nature of DNN systems and the increasing demand on model transparency and trustworthiness in the society. Government policies, i.e., the EU's General Data Protection Regulation (GDPR)~\cite{goodman2017european}, allude to a future where the explainability aspects of deep networks will become a legal concern.

We hope this field guide has distilled the essential topics, related work, methods, and concerns associated with explainable deep learning for an initiate. A wide range of existing methods on deep learning explainability is introduced and organized by a novel categorization scheme, depicting the field clearly and straightforwardly. Topics closely associated with DNN explainability, including model learning mechanism, model debugging, adversarial attack and defense, and model fairness and bias, are reviewed as related work. A discussion on user-oriented explanation designing and future trends of this field is provided at the end of this survey, shedding light on potential directions on model explainability. Given the countless papers in this field and the rapid development of explainable methods, we admit that we are unable to cover every paper or every aspect that belongs to this realm. We carefully designed hierarchical categories for the papers covered such that the skeleton of this field is visualized.

In the end, the important thing is to explain the right thing to the right person in the right way at the right time.\footnote{Paraphrased from Dr. Silja Renooij at BENELEARN2019} We are excited to continue to observe how the field evolves to deliver the appropriate explanation to the right audience who need it the most. We hope the numerous solutions actively being explored will lead to the fairer, safer, and more confident use of deep learning across society.

\acks{We acknowledge project support 
from the Ohio Federal Research Network, 
the Multidisciplinary Research Program of the Department
of Defense (MURI N00014-00-1-0637), 
and the organizers and participants of the Schloss Dagstuhl $-$ Leibniz Center for Informatics Seminar 17192 on Human-Like Neural-Symbolic Computing for providing the environment to develop the ideas in this paper. 
Parts of this work was completed under a Fulbright-NSF Fellowship for Cyber Security and Critical Infrastructure. 
We would also like to thank Erdi \c{C}all{\i} and Pim Haselager for the helpful discussions and general support.
}


\vskip 0.2in
\bibliography{ref}
\bibliographystyle{apalike}

\end{document}